\documentclass[lettersize,journal,compsoc]{IEEEtran}
\usepackage{amsfonts}
\usepackage{amsmath}
\usepackage{booktabs}
\usepackage[breaklinks=true,colorlinks,citecolor=blue,linkcolor=red]{hyperref}
\usepackage{gensymb}
\usepackage{graphicx}
\usepackage{multirow}
\usepackage{tabularx}
\usepackage{pifont}
\usepackage{url}

\usepackage{tikz}
\usepackage{pgf, pgfsys, pgffor}
\usepackage{pgfplotstable}
\pgfplotsset{compat=1.15}

\ifCLASSOPTIONcompsoc
  \usepackage[nocompress]{cite}
\else
  \usepackage{cite}
\fi

\newcommand{\cmark}{\textcolor{green}{\ding{51}}}%
\newcommand{\xmark}{\textcolor{red}{\ding{55}}}%
\newcommand{\dwar}{$\downarrow$}
\newcommand{\upar}{$\uparrow$}

\newcommand{\rev}[1]{#1}

\def\modelname{CityDreamer4D}
\def\acknowledgement{This study is supported by the Ministry of Education, Singapore, under its MOE AcRF Tier 2 (MOET2EP20221- 0012, MOE-T2EP20223-0002), and under the RIE2020 Industry Alignment Fund – Industry Collaboration Projects (IAF-ICP) Funding Initiative, as well as cash and in-kind contribution from the industry partner(s). (Corresponding author: Ziwei Liu.)}
\def\affiliations{The authors are with S-Lab, Nanyang Technological University, Singapore 637335 (email: haozhe.xie@ntu.edu.sg; zhaoxi001@ntu.edu.sg; fangzhou.hong@ntu.edu.sg; ziwei.liu@ntu.edu.sg)}
\def\projectpage{Project page is available at \protect\url{https://haozhexie.com/project/city-dreamer-4d}}

\begin{document}

\newcolumntype{Y}{>{\centering\arraybackslash}X}

\title{Compositional Generative Model of\\ Unbounded 4D Cities}

\author{
Haozhe Xie, Zhaoxi Chen, Fangzhou Hong, and Ziwei Liu
\ifCLASSOPTIONcompsoc
  \IEEEcompsocitemizethanks{%
    \IEEEcompsocthanksitem \acknowledgement%
    \IEEEcompsocthanksitem \affiliations%
    \IEEEcompsocthanksitem \projectpage%
  }
\else
  \thanks{\acknowledgement}%
  \thanks{\affiliations}%
  \thanks{\projectpage}%
\fi
}


\IEEEpubid{}
\IEEEtitleabstractindextext{
\begin{abstract}
3D scene generation has garnered growing attention in recent years and has made significant progress.
Generating 4D cities is more challenging than 3D scenes due to the presence of structurally complex, visually diverse objects like buildings and vehicles, and heightened human sensitivity to distortions in urban environments.
To tackle these issues, we propose \textbf{\modelname}, a compositional generative model specifically tailored for generating unbounded 4D cities.
Our main insights are 
\textbf{1)} 4D city generation should separate dynamic objects (\textit{e.g.}, vehicles) from static scenes (\textit{e.g.}, buildings and roads), and 
\textbf{2)} all objects in the 4D scene should be composed of different types of neural fields for buildings, vehicles, and background stuff.
Specifically, we propose Traffic Scenario Generator and Unbounded Layout Generator to produce dynamic traffic scenarios and static city layouts using a highly compact BEV representation.
Objects in 4D cities are generated by combining stuff-oriented and instance-oriented neural fields for background stuff, buildings, and vehicles. 
To suit the distinct characteristics of background stuff and instances, the neural fields employ customized generative hash grids and periodic positional embeddings as scene parameterizations.
Furthermore, we offer a comprehensive suite of datasets for city generation, including OSM, GoogleEarth, and CityTopia.
The OSM dataset provides a variety of real-world city layouts, while the Google Earth and CityTopia datasets deliver large-scale, high-quality city imagery complete with 3D instance annotations.
Leveraging its compositional design, \modelname ~supports a range of downstream applications, such as instance editing, city stylization, and urban simulation, while delivering state-of-the-art performance in generating realistic 4D cities.
\end{abstract}

}

\maketitle
\IEEEdisplaynontitleabstractindextext

\ifCLASSOPTIONcompsoc
  {\IEEEraisesectionheading{\section{Introduction}}}
\else
  {\section{Introduction}}
\fi

\IEEEPARstart{A}{mid} the rise of the metaverse, 3D and 4D asset generation has garnered significant attention. 
Notable progress has been made in generating 3D objects~\cite{DBLP:journals/ijcv/XieYZZS20,DBLP:conf/iclr/TangRZ0Z24,DBLP:preprint/arxiv/2409-12957}, avatars~\cite{DBLP:conf/iclr/Hong0LP023,DBLP:conf/nips/0009HMWYL23,DBLP:conf/cvpr/LiuZTSZLLL24}, and scenes~\cite{DBLP:journals/pami/ChenWL23,DBLP:journals/tog/WuLYSSWCLSLJ24,DBLP:preprint/arxiv/2406-06526}, 
as well as 4D objects~\cite{DBLP:conf/iclr/Jiang0GHY24,DBLP:preprint/arxiv/2406-10324} and avatars~\cite{DBLP:conf/icml/MaLJFSJ24,DBLP:preprint/arxiv/2406-04629,DBLP:preprint/arxiv/2407-06188}.
Cities, as one of the most essential assets, are widely used in diverse applications such as urban planning, environmental simulations, and game asset development.
Therefore, the challenge of making 3D/4D city development accessible to a wider audience, including artists, researchers, and players, becomes both significant and impactful.

In recent years, notable advancements have been made in scene generation.
Video-based methods~\cite{DBLP:conf/iccv/LiuM0SJK21,DBLP:conf/eccv/LiWSK22,DBLP:conf/siggraph/Deng0LGSW24} generate 3D scenes by producing videos conditioned on input images, but they cannot guarantee temporal consistency.
Outpainting-based methods~\cite{DBLP:conf/cvpr/LiangYLLXC24,DBLP:preprint/arxiv/2404-07199,DBLP:preprint/arxiv/2406-09394} generate 3D scenes through continuous outpainting on RGB and depth images, but they lack a compact scene representation, resulting in scenes that are typically small in scale.
PCG-based methods~\cite{DBLP:preprint/arxiv/2403-15698,DBLP:preprint/arxiv/abs-2406-04983,DBLP:preprint/arxiv/2407-17572} create unbounded cities by integrating large language models (LLMs) with procedural content generation (PCG), but the diversity of the generated cities is constrained by the 3D assets employed.
3D-aware-GAN-based methods, represented by GANCraft~\cite{DBLP:conf/iccv/HaoMB021} and SceneDreamer~\cite{DBLP:journals/pami/ChenWL23}, use volumetric neural rendering to generate images within a 3D scene, leveraging 3D coordinates and corresponding semantic labels.
These methods show promising results in generating 3D natural scenes by leveraging pseudo-ground-truth images generated by SPADE~\cite{DBLP:conf/cvpr/Park0WZ19}. 
InfiniCity~\cite{DBLP:conf/iccv/LinLMCS0T23} follows a similar pipeline for 3D city generation but it is more complex than 3D natural scenes due to the greater appearance variation in buildings and vehicles, unlike the relatively consistent appearance of objects with the same semantic label in natural scenes.
This variation reduces the quality of generated buildings and vehicles when all instances within their respective classes are assigned the same semantic label.
Generating 4D scenes poses greater challenges than 3D scenes, as existing methods~\cite{DBLP:conf/cvpr/BahmaniSRWGWTPT24,DBLP:conf/cvpr/ZhengLNLHM24,DBLP:preprint/arxiv/2403-16993,DBLP:conf/nips/YuWZMSCJTL24} either fail to ensure temporal consistency or are confined to tiny scales.

To address these problems, we propose \modelname, a compositional generative model designed for unbounded 4D cities.
As shown in Fig.~\ref{fig:overview}, the unbounded 4D city generation framework separates dynamic objects from static scenes.
Static scenes are defined by the city layout from Unbounded Layout Generator, arranging elements like roads, highways, vegetation, and buildings, with the capability to extrapolate to unbounded areas.
Dynamic objects, such as vehicles, are defined by traffic scenarios generated by Traffic Scenario Generator, which determines their spatial positioning on high-definition (HD) maps derived from city layouts.
Unlike existing methods that use a single module for all objects, \modelname~divides the generation process into three distinct modules: Building Instance Generator for buildings, Vehicle Instance Generator for vehicles, and City Background Generator for background stuff.
These generators leverage a highly compact bird's-eye-view (BEV) scene representation to ensure efficiency and scalability.
The scene parameterization is designed to address the unique characteristics of background stuff and instances: background stuff often features similar appearances with irregular textures, while buildings and vehicles display diverse appearances with regular periodic patterns. 
To handle these variations, we use generative hash grids for the background and apply periodic positional encodings to each instance. 
We also place buildings in an object-centric coordinate space and vehicles in an object-canonical coordinate space, using specialized methods designed to capture their compact shapes.
Compositor combines the rendered background stuff with the building and vehicle instances to create a unified image.

To improve the realism of our generated cities, we construct a suite of datasets, including OSM, GoogleEarth, and CityTopia.
The OSM dataset, sourced from OpenStreetMap~\cite{HAOZHE:link/OpenStreetMap}, includes semantic maps and height fields for 80 cities worldwide, covering over 6,000 km\textsuperscript{2}. 
The semantic maps indicate the locations of roads, buildings, urban greenery, and water bodies, while the height fields primarily represent building heights.
The GoogleEarth dataset is a real-world dataset collected using Google Earth Studio~\cite{HAOZHE:link/GoogleEarth}, featuring 400 drone-view orbit trajectories over New York City.
It includes 24,000 real-world city images, with 3D semantic annotations for all classes and 3D instance annotations for buildings.
The CityTopia dataset is a high-quality synthetic dataset spanning 11 cities generated with 3D assets from the Unreal Engine 5 City Sample project~\cite{HAOZHE:link/CitySample}.
It offers 37,500 high-fidelity street-view and drone-view images, featuring precise 2D and 3D semantic annotations for all classes, along with 3D instance annotations for buildings and vehicles.

The contributions are summarized as follows:
\begin{itemize}
\item We propose \modelname, the first generative model for unbounded 4D cities that disentangles dynamic objects from static scenes and enables instance editing, city stylization, and urban simulation.
\item We introduce stuff-oriented and instance-oriented neural fields to generate background stuff and instances (buildings and vehicles) in 4D scenes, effectively capturing their diversity.
\item We create comprehensive datasets for city generation, using OSM for realistic layouts and Google Earth and CityTopia for detailed city visuals with 3D semantic and instance annotations.
\item The proposed \modelname ~demonstrates superior capability in generating unbounded, diverse 4D cities and enables instance-level editing within them.
\end{itemize}

A preliminary version of this work, named CityDreamer, has been published in CVPR 2024~\cite{DBLP:conf/cvpr/XieCHL24}.
We make several extensions in this work compared to the preliminary version.
\textbf{1)} We evolve CityDreamer into \modelname, enabling 4D city generation through Traffic Scenario Generator and Vehicle Instance Generator, effectively separating dynamic objects from static scenes.
\textbf{2)} We enhance the highly compact BEV representation by incorporating an additional bottom-up height map, enabling the representation of hollow structures in cities, such as highways.
\textbf{3)} We propose Traffic Scenario Generator, which creates HD maps from city layouts to produce realistic traffic scenarios with vehicles in unbounded cities.
\textbf{4)} We introduce Vehicle Instance Generator, designed to generate vehicle instances within cities using a novel scene parameterization method grounded on the canonical feature space.
\textbf{5)} We build the CityTopia dataset, offering nearly 40k high-quality street-view and drone-view images with both 2D and 3D semantic and instance annotations.

\section{Related Works}

\subsection{3D-aware GANs}

Building on the recent success of 2D GANs~\cite{DBLP:journals/pami/KarrasLA21,DBLP:journals/pami/MelnikMMPAHRRR24}, various approaches have been introduced to generate 3D content using GANs as well.
The core idea is to represent the generated scenes using a 3D representation and apply rendering techniques to produce images from various viewpoints, enabling image-level adversarial learning~\cite{DBLP:conf/nips/GoodfellowPMXWOCB14}.
Early methods use explicit shapes like voxels~\cite{DBLP:conf/3dim/GadelhaMW17,DBLP:conf/iccv/HenzlerM019,DBLP:conf/iccv/Nguyen-PhuocLTR19}, meshes~\cite{DBLP:preprint/arxiv/1910-00287}, and 3D primitives~\cite{DBLP:conf/cvpr/LiaoSMG20} to render images from different viewpoints. 
However, their limited expressiveness and efficiency hinder the synthesis of complex scenes and photorealistic details. 
NeRF~\cite{DBLP:conf/eccv/MildenhallSTBRN20}, known for producing high-fidelity novel view synthesis, are introduced to 3D-aware generative models. 
Yet, the high computational cost of NeRF-based GANs restricts earlier attempts~\cite{DBLP:conf/nips/SchwarzLN020,DBLP:conf/cvpr/ChanMK0W21,DBLP:conf/iccv/DeVries0STS21,DBLP:conf/nips/XuPLD21} from generating high-quality images. 
To address this, many follow-up works~\cite{DBLP:conf/cvpr/Niemeyer021,DBLP:conf/iclr/GuL0T22,DBLP:conf/cvpr/XueLSL22,DBLP:conf/cvpr/Or-ElLSSPK22,DBLP:conf/cvpr/ChanLCNPMGGTKKW22} avoid rendering NeRFs at high resolution by applying 2D super-resolution on low-resolution feature maps, though this compromises 3D consistency.
To maintain strict 3D consistency, newer approaches shift to sparser 3D representations, like sparse voxels~\cite{DBLP:conf/nips/SchwarzSNL022}, radiance manifolds~\cite{DBLP:conf/cvpr/DengYX022}, and multi-plane images~\cite{DBLP:conf/eccv/ZhaoMGRSC22}, enabling direct high-resolution rendering.
Nevertheless, most of these methods are trained on curated datasets for bounded scenes, such as human faces~\cite{DBLP:conf/cvpr/KarrasLA19,DBLP:conf/cvpr/Yang0WHSYC20}, human bodies~\cite{DBLP:journals/pami/IonescuPOS14,DBLP:conf/eccv/CaiRZLYWFGYPHZL22}, and objects~\cite{DBLP:conf/cvpr/WuZFWRPWYWQLL23,DBLP:conf/cvpr/DeitkeSSWMVSEKF23}.

\subsection{3D Scene Generation}

Unlike advanced 2D generative models that mainly focus on individual categories or familiar objects, generating scene-level content is more challenging due to the vast diversity and complexity of scenes~\cite{DBLP:preprint/arxiv/2505-05474}.
Earlier methods~\cite{DBLP:conf/iccv/LiuM0SJK21,DBLP:conf/eccv/LiWSK22} generate scenes by synthesizing videos, but they lack 3D awareness and fail to ensure 3D consistency. 
Semantic image synthesis approaches~\cite{DBLP:conf/iccv/HaoMB021,DBLP:journals/pami/ZhanYWZLLKTX23} have shown promising results in generating scene-level content by conditioning on pixel-wise dense correspondences, like semantic segmentation or depth maps. 
Several techniques~\cite{DBLP:conf/cvpr/LiangYLLXC24,DBLP:preprint/arxiv/2404-07199,DBLP:preprint/arxiv/2406-09394} generate 3D scenes by performing inpainting and outpainting on RGB images or feature maps, though most can only interpolate or extrapolate a limited distance from the input views and lack true generative capabilities. 
Significant research has investigated procedural content generation (PCG) for creating natural~\cite{DBLP:conf/cvpr/RaistrickLMMWZK23,DBLP:preprint/arxiv/2403-15698}, indoor~\cite{DBLP:journals/tog/LiPXCKSTCCZ19,DBLP:conf/iccv/FuC0ZWLZSJZ021,DBLP:conf/cvpr/RaistrickMKY0HW24}, and urban scenes~\cite{DBLP:conf/corl/DosovitskiyRCLK17,DBLP:preprint/arxiv/1909-11512,DBLP:preprint/arxiv/2407-17572,DBLP:preprint/arxiv/2412-07660}, but the diversity of the generated scenes is limited by the 3D assets used.
Recent methods~\cite{DBLP:conf/iccv/LinLMCS0T23,DBLP:journals/pami/ChenWL23,DBLP:preprint/arxiv/2406-06526} achieve 3D-consistent scenes at an infinite scale through unbounded layout extrapolation. 
Other approaches~\cite{DBLP:journals/pami/JiangKS16,DBLP:conf/nips/PaschalidouKSKG21,DBLP:journals/pami/GaoSMLGY23} focus on indoor scene synthesis, relying on costly 3D datasets~\cite{DBLP:conf/cvpr/DaiCSHFN17,DBLP:preprint/arxiv/1906-05797} or CAD object retrieval~\cite{DBLP:conf/cvpr/SongYZCSF17,DBLP:journals/ijcv/FuJGGZMT21,DBLP:conf/corl/DaiWJWGZWF24}.

\begin{figure*}[!t]
  \includegraphics[width=\linewidth]{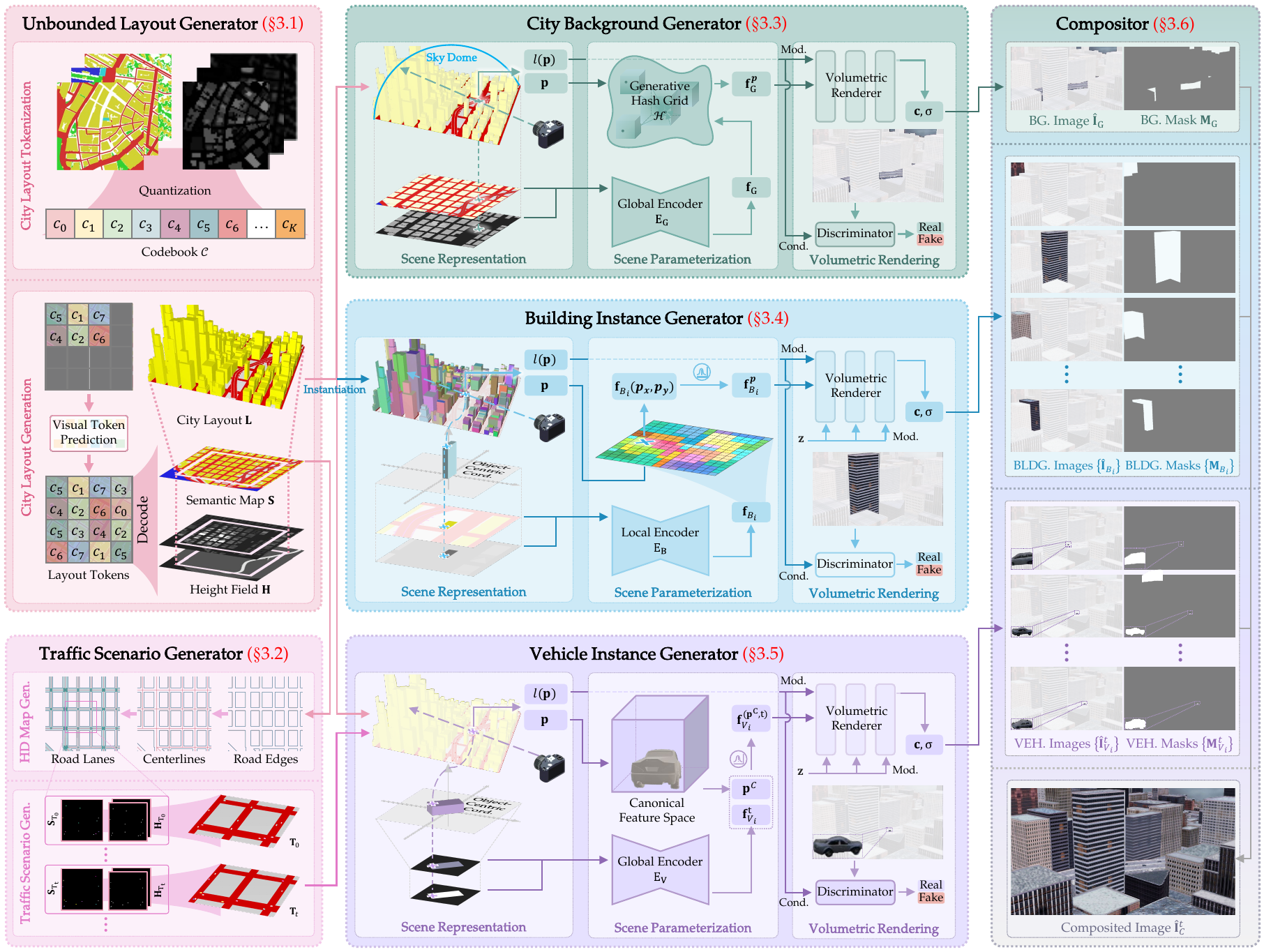}
  \caption{\textbf{Overview of \modelname.} 4D city generation comprises static and dynamic scenes, conditioned on city layout $\mathbf{L}$ and time-varying traffic scenario $\mathbf{T}_t$, generated by the Unbounded Layout and Traffic Scenario Generators, respectively. City Background Generator uses $\mathbf{L}$ to create background images $\mathbf{\hat{I}}_{G}$ for stuff like roads, vegetation, and the sky, while Building Instance Generator renders the buildings $\{\mathbf{\hat{I}}_{B_i}\}$ within the city. Using $\mathbf{T}_t$, Vehicle Instance Generator generates vehicles $\{\mathbf{\hat{I}}_{V_i}^t\}$ at time step $t$. Finally, Compositor combines the rendered background, buildings, and vehicles into a unified and coherent image $\mathbf{\hat{I}}_{C}^t$. ``Gen.'', ``Mod.``, ``Cond.'', ``BG.'', ``BLDG.'', and ``VEH.'' denote ``Generation'', ``Modulation'', ``Condition'', ``Background'', ``Building'', and ``Vehicle'', respectively.}
  \label{fig:overview}
\end{figure*}

\subsection{4D Scene Generation}

In recent years, representations like D-NeRF~\cite{DBLP:conf/cvpr/PumarolaCPM21} and Deformable 3D Gaussians~\cite{DBLP:conf/cvpr/YangGZJ0024} have been proposed for 4D object and human generation.
However, 4D scene generation remains in its early stages, as existing representations are not designed for large-scale scene generation.
Mainstream approaches typically formulate it as 4D occupancy generation~\cite{DBLP:preprint/arxiv/2410-10429,DBLP:preprint/arxiv/2410-18084} and distillation from video diffusion~\cite{DBLP:conf/cvpr/BahmaniSRWGWTPT24,DBLP:conf/cvpr/ZhengLNLHM24,DBLP:preprint/arxiv/2403-16993,DBLP:conf/nips/YuWZMSCJTL24}. 
However, these methods lack compact representations, restricting the scale of the generated scenes.

\section{Method}

As illustrated in Figure~\ref{fig:overview}, \modelname ~decouples unbounded 4D city generation into static scene generation and dynamic object generation.
For static scene generation, Unbounded Layout Generator (Section~\ref{sec:unbounded-layout-generator}) creates an arbitrarily large city layout $\mathbf{L}$. 
City Background Generator (Section~\ref{sec:city-bg-generator}) then produces the background image $\mathbf{\hat{I}}_\text{G}$ along with its corresponding mask $\mathbf{M}_\text{G}$. 
Following this, Building Instance Generator (Section~\ref{sec:building-ins-generator}) generates images for building instances $\{\mathbf{\hat{I}}_{B_i}\}_{i=1}^{n_\text{B}}$ and their respective masks $\{\mathbf{M}_{B_i}\}_{i=1}^{n_\text{B}}$, where $n_\text{B}$ is the number of building instances.
For dynamic object generation, the traffic generator (Section~\ref{sec:traffic-generator}) first creates the traffic scenario $\mathbf{T}_t$ for time step $t$. 
Then, Vehicle Instance Generator (Section~\ref{sec:vehicle-ins-generator}) produces images for vehicle instances $\{\mathbf{\hat{I}}_{\text{V}_i}^t\}_{i=1}^{n_\text{V}}$ and their corresponding masks $\{\mathbf{M}_{\text{V}_i}^t\}_{i=1}^{n_\text{V}}$ based on $\mathbf{T}_t$, where $n_\text{V}$ denotes the number of vehicles.
Finally, Compositor (Section~\ref{sec:compositor}) merges the rendered background, building instances, and vehicle instances into a cohesive image $\mathbf{I}_{\rm C}^t$ for time step $t$.

\subsection{Unbounded Layout Generator}
\label{sec:unbounded-layout-generator}

\noindent \textbf{City Layout Representation.}
The city layout defines the locations of static 3D objects within the city, which are grouped into categories such as roads, highways, buildings, vegetation, water areas, and others.
Additionally, a null class is included to represent empty spaces in the 3D volume. 
The city layout in \modelname, represented as a 3D volume $\mathbf{L}$, is constructed by extruding pixels from the semantic map $\mathbf{S}_L$ according to their corresponding values in the height field $\mathbf{H}_L$ = $\left\{\mathbf{H}_L^\textrm{BU}, \mathbf{H}_L^\textrm{TD}\right\}$, where $\mathbf{H}_L^\textrm{BU}$ and $\mathbf{H}_L^\textrm{TD}$ represent the bottom-up heights and the top-down heights, respectively.
Specifically, the value of $\mathbf{L}$ at $(i, j, k)$ is defined as
\begin{equation}
  \mathbf{L}{(i, j, k)} = 
  \begin{cases}
    \mathbf{S}_L{(i, j)} & \textrm{if}~\mathbf{H}_L^\textrm{BU}{(i, j)} \leq k \leq \mathbf{H}_L^\textrm{TD}{(i, j)} \\
    0          & \textrm{otherwise}
  \end{cases}
  \label{eq:layout-gen}
\end{equation}
where $0$ denotes empty spaces in the 3D volumes.

\noindent \textbf{City Layout Generation.}
Obtaining unbounded city layouts is translated into generating extendable semantic maps and height fields.
To achieve this, we design Unbounded Layout Generator based on MaskGIT~\cite{DBLP:conf/cvpr/ChangZJLF22}, which naturally supports inpainting and extrapolation.
Specifically, we leverage VQVAE~\cite{DBLP:conf/nips/OordVK17} to tokenize patches of semantic maps and height fields, encoding them into a discrete latent space with a codebook $\mathcal{C} = \left\{c_k \mid c_k \in \mathbb{R}^{d_C}\right\}_{k=1}^{d_K}$.
During inference, the layout tokens are generated autoregressively, and the VQVAE decoder reconstructs a pair of semantic map $\mathbf{S}_L$ and height field $\mathbf{H}_L$. 
Since VQVAE produces fixed-size outputs, we perform image extrapolation to create arbitrarily large layouts. 
This involves using a sliding window with a 25\% overlap to iteratively predict local layout tokens at each step.

\noindent \textbf{Loss Functions.}
The VQVAE handles the generation of the height field and semantic map as separate tasks, optimized with L1 Loss and Cross-Entropy Loss $\mathcal{E}$, respectively. 
To enhance the sharpness of the height field near building edges, we incorporate an additional Smoothness Loss $\mathcal{S}$~\cite{DBLP:conf/aaai/MeisterH018}
\begin{equation}
  \ell_{\rm VQ} = \lambda_{\rm R} \lVert \mathbf{\hat{H}}_L^p - \mathbf{H}_L^p \rVert
                + \lambda_{\rm S} \mathcal{S}(\mathbf{\hat{H}}_L^p, \mathbf{H}_L^p)
                + \lambda_{\rm E} \mathcal{E}(\mathbf{\hat{S}}_L^p, \mathbf{S}_L^p)
\end{equation}
where $\mathbf{\hat{H}}_L^p$ and $\mathbf{\hat{S}}_L^p$ denote the generated height field and semantic map patches, respectively.
$\mathbf{H}_L^p$ and $\mathbf{S}_L^p$ are the corresponding ground truth.
MaskGIT's autoregressive transformer is optimized with a reweighted ELBO loss~\cite{DBLP:conf/eccv/Bond-TaylorH0BW22}.

\subsection{Traffic Scenario Generator}
\label{sec:traffic-generator}

\noindent \textbf{Traffic Scenario Representation.}
The city layout $\mathbf{L}$ defines the static elements of the unbounded city, while the dynamic aspects are captured by the traffic scenario, represented as $\mathcal{T} = \left\{\mathbf{T}_t\right\}_{t=1}^{n_T}$, where $n_T$ represents the number of frames. 
Similar to the city layout $\mathbf{L}$, $\mathbf{T}_t$ is likewise derived from the semantic map $\mathbf{S}_{T_t}$ and the height field $\mathbf{H}_{T_t} = \left\{\mathbf{H}_{T_t}^\textrm{BU}, \mathbf{H}_{T_t}^\textrm{TD}\right\}$, where the semantic map specifies the positions of dynamic objects, and the height field defines their elevations.
Specifically, the value of $\mathbf{T}_t$ at $(i, j, k)$ is
\begin{equation}
  \mathbf{T}_t{(i, j, k)} = 
  \begin{cases}
    \mathbf{S}_{T_t}{(i, j)} & \textrm{if}~\mathbf{H}_{T_t}^\textrm{BU}{(i, j)} \leq k \leq \mathbf{H}_{T_t}^\textrm{TD}{(i, j)} \\
    0          & \textrm{otherwise}
  \end{cases}
  \label{eq:traffic-gen}
\end{equation}
where $0$ denotes empty spaces in the 3D volumes.

\noindent \textbf{Traffic Scenario Generation.}
The generation of traffic scenario $\mathcal{T}$ is conceptualized as the frame-by-frame production of semantic maps $\mathbf{S}_T = \left\{\mathbf{S}_{T_t}\right\}_{t=1}^{n^T}$ and height fields $\mathbf{H}_T = \left\{\mathbf{H}_{T_t}\right\}_{t=1}^{n^T}$.
To guarantee realistic and continuous placement of dynamic objects, a high-definition (HD) map is derived from the city layout $\mathbf{L}$.
Unlike the city layout, which only specifies the positions of roads and highways, the HD map includes details about lanes, intersections, and traffic signals.
Using the generated HD map, an off-the-shelf model~\cite{DBLP:conf/icra/FengLPTZ23} determines the per-frame bounding boxes of dynamic objects.
The corresponding semantic map and height field are generated based on the bounding boxes.

\noindent \textbf{HD Map Generation.}
In HD maps, we adopt the entity definitions from the Waymo Motion dataset~\cite{DBLP:conf/iccv/EttingerCCLZPCS21}, which include road edges, road lanes, road lines, stop signs, and traffic lights.

\noindent \textit{Road Edges}, representing the boundaries of roads, are generated by applying Canny edge detection~\cite{DBLP:journals/pami/Canny86a} to $\mathbf{S}_L$ and converting the continuous edges into a graph structure using vectorization, which involves detecting corner points and connecting them sequentially.

\noindent \textit{Road Lanes}, representing the centerlines of lanes where vehicles can travel, are derived by skeletonizing~\cite{DBLP:journals/cacm/ZhangS84} $\mathbf{S}_L$ to extract road structures and identifying intersections where multiple edges connect.
The image is then converted into road centerline graphs using graph-based traversal. 
The number and positions of the lanes are determined based on road width, and lanes at intersections are connected using B\'ezier curves.

\noindent \textit{Road Lines}, such as solid single white or solid double yellow, are generated according to the positions and attributes of the road lanes.

\noindent \textit{Stop Signs} and \textit{Traffic Lights} are positioned at the intersections, where multiple road lanes converge.

\subsection{City Background Generator}
\label{sec:city-bg-generator}

\noindent \textbf{Scene Representation.}
Following SceneDreamer~\cite{DBLP:journals/pami/ChenWL23}, we adopt a bird's-eye-view (BEV) representation for its efficiency and expressiveness, particularly suited for unbounded scenes.
Unlike GANCraft~\cite{DBLP:conf/iccv/HaoMB021} and InfiniCity~\cite{DBLP:conf/iccv/LinLMCS0T23}, which parameterize features at voxel corners, our BEV representation uses a feature-free 3D volume constructed from a height field and a semantic map, as described in Equation~\ref{eq:layout-gen}.
Specifically, we extract a local window of resolution $N^H_G \times N^W_G \times N^D_G$ from the city layout $\mathbf{L}$. 
This local window $\mathbf{L}^G$ is generated using the corresponding height field $\mathbf{H}_L^G$ and semantic map $\mathbf{S}_L^G$.

\noindent \textbf{Scene Parameterization.}
To achieve generalizable 3D representation learning across various scenes and align content with 3D semantics, it is necessary to parameterize the scene representation into a latent space, making adversarial learning easier.
For background stuff, we adopt the generative neural hash grid~\cite{DBLP:journals/pami/ChenWL23} to learn generalizable features across scenes by modeling the hyperspace beyond 3D space.
Specifically, we first encode the local scene $(\mathbf{H}_L^G, \mathbf{S}_L^G)$ using the global encoder $\textrm{E}_\textrm{G}$ to produce the compact scene-level feature $\mathbf{f}_G \in \mathbb{R}^{d_G}$.
\begin{equation}
  \mathbf{f}_G = \textrm{E}_\textbf{G}(\mathbf{H}_L^G, \mathbf{S}_L^G)
  \label{eq:global-encoder}
\end{equation}
Using a learnable neural hash function $\mathcal{H}$, the indexed feature $\mathbf{f}_G^{\mathbf{p}}$ at the 3D position $\mathbf{p} \in \mathbb{R}^3$ is derived by mapping $\mathbf{p}$ and $\mathbf{f}_G$ into a hyperspace, specifically $\mathbb{R}^{3 + d_G} \rightarrow \mathbb{R}^{N_G^C}$.
\begin{equation}
  \mathbf{f}_G^{\mathbf{p}} =
  \mathcal{H}(\mathbf{p}, \mathbf{f}_G) = 
      \Big(\bigoplus^{d_G}_{i=1}f_G^i\pi^i \bigoplus^3_{j=1}p^j\pi^j \Big) \mod N_E
  \label{eq:hashgrid}
\end{equation}
where $\oplus$ represents the bitwise XOR operation, while $\pi^i$ and $\pi^j$ are distinct large prime numbers.
To capture multi-scale features, we construct $N_H^L$ levels of multi-resolution hash grids. 
Each level contains up to $N_E$ entries, with $N_G^C$ denoting the number of channels in each feature vector.

\noindent \textbf{Volumetric Rendering.}
In the perspective camera model, every pixel in the image is associated with a camera ray $\mathbf{r}(t) = \mathbf{o} + t\mathbf{v}$, which originates at the projection center $\mathbf{o}$ and extends along the direction $\mathbf{v}$.
The pixel value $C(\mathbf{r})$ is then computed as an integral along this ray.
\begin{equation}
  C({\mathbf{r}}) =\int^{\infty}_{0}
  A(t)
  \mathbf{c}(\mathbf{f}_G^{\mathbf{r}(t)}, l(\mathbf{r}(t)))
  \boldsymbol{\sigma}(\mathbf{f}_G^{\mathbf{r}(t)}) dt
\end{equation}
where $A(t) = \exp\left(-\int^t_0 \sigma(\mathbf{f}_G^{\mathbf{r}(s)}) , ds\right)$ represents the accumulated transmittance.
$l(\mathbf{p})$ denotes the semantic label at the 3D position $\mathbf{p}$.
The symbols $\mathbf{c}$ and $\boldsymbol{\sigma}$ correspond to the color and volume density, respectively.

\noindent \textbf{Loss Function.}
City Background Generator is optimized with a hybrid objective that combines reconstruction loss and adversarial loss.  
In particular, it uses an L1 loss, a perceptual loss $\mathcal{P}$~\cite{DBLP:conf/eccv/JohnsonAF16}, and a GAN loss $\mathcal{G}$~\cite{DBLP:arxiv/LimY17} as part of this objective.
\begin{equation}
  \ell_G = \lambda_G^{\rm L1} \lVert\mathbf{\hat{I}}_G - \mathbf{I}_G\rVert
         + \lambda_G^{\rm P} \mathcal{P}(\mathbf{\hat{I}}_G, \mathbf{I}_G)
         + \lambda_G^{\rm G} \mathcal{G}(\mathbf{\hat{I}}_G, \mathbf{S}_G)
\end{equation}
where $\mathbf{I}_G$ represents the ground truth background image, while $\mathbf{S}_G$ corresponds to the perspective-view semantic map obtained by accumulating semantic labels sampled from $\mathbf{L}^G$ along each ray.  
The weights for the three losses are denoted by $\lambda_G^{\rm L1}$, $\lambda_G^{\rm P}$, and $\lambda_G^{\rm G}$.
Note that $\ell_G$ is only applied to pixels whose semantic labels are classified as background stuff.

\subsection{Building Instance Generator}
\label{sec:building-ins-generator}

\noindent \textbf{Scene Representation.}
Building Instance Generator also employs the BEV scene representation. 
It extracts a local window $\mathbf{L}^{B_i}$ from the city layout $\mathbf{L}$ with dimensions $N_B^H \times N_B^W \times N_B^D$.
This window is centered around the 2D coordinates $(c_x^{B_i}, c_y^{B_i})$ of the building instance $B_i$. 
The height field and semantic map used to construct $\mathbf{L}^{B_i}$ are represented as $\mathbf{H}_L^{B_i}$ and $\mathbf{S}_L^{B_i}$, respectively.
Since all buildings share the same semantic label in $\mathbf{S}_L$, we perform building instantiation by detecting connected components.
Notably, real-world building facades and roofs exhibit distinct visual distributions. 
To capture this, we assign different semantic labels to the facade and roof of each building instance $\rm B_i$ in $\mathbf{L}^{B_i}$, with the roof assigned to the top-most voxel layer.
All other building instances are excluded from $\mathbf{L}^{B_i}$ by assigning them a value of $0$.

\noindent \textbf{Scene Parameterization.}
Unlike City Background Generator, Building Instance Generator employs a distinct scene parameterization, encoding the local scene $(\mathbf{H}_L^{B_i}, \mathbf{S}_L^{B_i})$ with $\textrm{E}_\textrm{B}$ to produce pixel-level features $\mathbf{f}_{B_i}$ of resolution $N_B^H \times N_B^W \times N_B^C$.
\begin{equation}
  \mathbf{f}_{B_i} = \textrm{E}_\textrm{B}(\mathbf{H}_L^{B_i}, \mathbf{S}_L^{\rm B_i})
  \label{eq:bldg-local-encoder}
\end{equation}
For a 3D position $\mathbf{p} = (p_x, p_y, p_z)$, the corresponding feature $\mathbf{f}_{\rm B_i}^{\mathbf{p}}$ is obtained as
\begin{equation}
  \mathbf{f}_{B_i}^{\mathbf{p}} = \mathcal{O}({\rm Concat}(\mathbf{f}_{B_i}(p_x, p_y), p_z))
\end{equation}
where $\rm Concat(\cdot)$ denotes the concatenation operation.
$\mathbf{f}_{B_i}(p_x, p_y) \in \mathbb{R}^{N_B^C}$ represents the feature vector corresponding to the coordinates $(p_x, p_y)$.
$\mathcal{O}(\cdot)$ refers to the positional encoding function adopted in the standard NeRF~\cite{DBLP:conf/eccv/MildenhallSTBRN20}.
\begin{equation}
  \mathcal{O}(\mathbf{x}) = \{\sin(2^i \pi \mathbf{x}), \cos(2^i \pi \mathbf{x})\}_{i=0}^{N_P^L - 1}
  \label{eq:sincos}
\end{equation}
Note that $\mathcal{O}(\cdot)$ is applied separately to each element of the feature $\mathbf{x}$, with the values normalized to the range $[-1, 1]$.

\noindent \textbf{Volumetric Rendering.}
Unlike the volumetric rendering approach used in City Background Generator, Building Instance Generator incorporates a style code $\mathbf{z}$ to capture the variability in building appearances. 
The pixel value $C(\mathbf{r})$ is computed through an integration process.
\begin{equation}
  C({\mathbf{r}}) =\int^{\infty}_{0}
  A(t)
  \mathbf{c}(\mathbf{f}_{B_i}^{\mathbf{r}(t)}, \mathbf{z}, l(\mathbf{r}(t)))
  \boldsymbol{\sigma}(\mathbf{f}_{B_i}^{\mathbf{r}(t)}) dt
  \label{eq:vol-render-z}
\end{equation}
where $\mathbf{r}(t) = \mathbf{o} + t\mathbf{v} - \left[c_x^{B_i}, c_y^{B_i}, 0\right]^\textrm{T}$, which is employed to center the buildings within their local coordinate system.

\noindent \textbf{Loss Function.}
The training of Building Instance Generator relies solely on the GAN loss $\mathcal{G}$, formulated as
\begin{equation}
  \ell_B = \mathcal{G}(\mathbf{\hat{I}}_{B_i}, \mathbf{S}_{B_i})
\end{equation}
where $\mathbf{S}_{B_i}$ represents the semantic map of the building instance $\rm B_i$ in perspective view, generated similarly to $\mathbf{S}_G$.
Note that $\ell_B$ is only applied to pixels with semantic labels corresponding to the building instance.

\begin{table*}
  \setlength\tabcolsep{4pt}
  \setlength\extrarowheight{2pt}
  \caption{\textbf{Comparison of Statistics and Properties: GoogleEarth, CityTopia, and Previous Datasets.} Only annotated images are counted. ``Ext.'' stands for ``Extendable'', indicating whether the dataset can be easily expanded following the current data generation pipeline. ``3DM.'', ``Sem.'', and ``Inst.'' refer to ``3D Model'', ``Semantic'', and ``Instance'', respectively.}
  \label{tab:comp-dataset}
  \begin{tabularx}{\linewidth}{l|cccccc|Yc|Yc|cccc}
    \toprule
    \multirow{2}{*}{Dataset} & 
    \#Images &
    \multirow{2}{*}{\#Cities} &
    Area & 
    \multirow{2}{*}{Source} & 
    \multirow{2}{*}{Ext.} & 
    \multirow{2}{*}{3DM.} & 
    \multicolumn{2}{c|}{Lighting} & 
    \multicolumn{2}{c|}{View Type} & 
    \multicolumn{4}{c}{Dense Annotations} \\
    \cline{8-9} \cline{10-11} \cline{12-15}
    & 
    ($\times 10^3$) & 
    & 
    (km\textsuperscript{2}) & 
    & 
    & 
    & 
    Day         & Night    & 
    Street      & Aerial   & 
    2D Sem.     & 2D Inst. & 3D Sem. & 3D Inst.\\ 
    \midrule
    KITTI~\cite{DBLP:conf/cvpr/GeigerLU12} &
    0.2         & 1        & -        & Real      & \xmark & \xmark &
    \cmark      & \xmark   & \cmark   & \xmark    & 
    \cmark      & \cmark   & \xmark   & \xmark  \\
    Cityscapes~\cite{DBLP:conf/cvpr/CordtsORREBFRS16} &
    25          & 50       & -        & Real      & \xmark & \xmark &
    \cmark      & \xmark   & \cmark   & \xmark    & 
    \cmark      & \cmark   & \xmark   & \xmark  \\
    AeroScapes~\cite{DBLP:conf/wacv/NigamHR18} &
    3.2         & -        & -        & Real      & \xmark & \xmark &
    \cmark      & \xmark   & \xmark   & \cmark    & 
    \cmark      & \xmark   & \xmark   & \xmark  \\
    nuScenes~\cite{DBLP:conf/cvpr/CaesarBLVLXKPBB20} &
    93          & 2        & -        & Real      & \xmark & \xmark & 
    \cmark      & \xmark   & \cmark   & \xmark    & 
    \cmark      & \cmark   & \xmark   & \xmark  \\
    \midrule
    GTA-V~\cite{DBLP:conf/eccv/RichterVRK16} &
    25          & -        & -        & Synthetic & \xmark & \xmark & 
    \cmark      & \xmark   & \cmark   & \xmark    & 
    \cmark      & \xmark   & \xmark   & \xmark    \\
    SYNTHIA~\cite{DBLP:conf/cvpr/RosSMVL16} &
    213         & 1        & -        & Synthetic & \xmark & \xmark & 
    \cmark      & \xmark   & \cmark   & \cmark    & 
    \cmark      & \cmark   & \xmark   & \xmark    \\
    VEIS~\cite{DBLP:conf/eccv/SalehASPA18} &
    61          & -        & -        & Synthetic & \xmark & \xmark & 
    \cmark      & \xmark   & \cmark   & \cmark    & 
    \cmark      & \cmark   & \xmark   & \xmark    \\
    MatrixCity~\cite{DBLP:conf/iccv/0002JXX0L023} & 
    519         & 2        & 28       & Synthetic & \xmark & \xmark &
    \cmark      & \cmark   & \cmark   & \cmark    & 
    \xmark      & \xmark   & \xmark   & \xmark  \\
    \midrule
    HoliCity~\cite{DBLP:preprint/arxiv/2008-03286} &
    6.3         & 1        & 20       & Real      & \xmark & CAD &
    \cmark      & \xmark   & \cmark   & \xmark    & 
    \cmark      & \cmark   & \xmark   & \xmark  \\
    KITTI-360~\cite{DBLP:journals/pami/LiaoXG23} & 
    78          & 1        & -        & Real      & \xmark & CAD &
    \cmark      & \xmark   & \cmark   & \xmark    & 
    \cmark      & \cmark   & \xmark   & \xmark  \\
    UrbanScene3D~\cite{DBLP:conf/eccv/LinLHYXH22} & 
    6.1$^\dag$  & -        & 3$^\dag$ & Real      & \xmark & Mesh &
    \cmark      & \xmark   & \xmark   & \cmark    & 
    \xmark      & \xmark   & \xmark   & \cmark  \\
    GoogleEarth &
    24          & 1        & 25       & Real      & \cmark & Voxel &
    \cmark      & \xmark   & \xmark   & \cmark    & 
    \cmark      & \cmark   & \cmark   & \cmark  \\
    \midrule
    CityTopia   &
    37.5        & 11       & 36       & Synthetic & \cmark & Voxel &
    \cmark      & \cmark   & \cmark   & \cmark    &
    \cmark      & \cmark   & \cmark   & \cmark    \\
    \bottomrule
    \multicolumn{15}{l}{$^\dag$ Only the real-world image subset is counted for this dataset.}
  \end{tabularx}
\end{table*}


\subsection{Vehical Instance Generator}
\label{sec:vehicle-ins-generator}

\noindent \textbf{Scene Representation.}
Vehicle Instance Generator, like Building Instance Generator, leverages the BEV scene representation. 
It extracts a local window $\mathbf{T}_t^{V_i}$ from the traffic scenario $\mathbf{T}_t$, with dimensions $N_V^H \times N_V^W \times N_V^D$, to generate the vehicle instances within the scene.
This window is centered around the 2D coordinates $(c_x^{V_i}, c_y^{V_i})$ of the vehicle instance $V_i$. 
The height field and semantic map used to construct $\mathbf{T}_t^{V_i}$ are represented as $\mathbf{H}_{T_t}^{V_i}$ and $\mathbf{S}_{T_t}^{V_i}$, respectively.
Unlike buildings, vehicle instances are instantiated during the generation of the traffic scenario.
Instances other than $V_i$ are removed from $\mathbf{T}_t^{V_i}$ by assigning them a value of $0$.

\noindent \textbf{Scene Parameterization.}
Compared to building instances, vehicle instances demonstrate greater structural regularity, closely tied to their relative positions. 
For instance, within the same vehicle, the front, rear, and body exhibit distinct appearances, yet these structural features remain consistent across different vehicles. 
Building on this observation, we propose a scene parameterization method based on the canonical feature space. 
Given a 3D position $\mathbf{p} = (p_x, p_y, p_z)$, the canonicalized point $\mathbf{p}^C$ is
\begin{equation}
  \mathbf{p}^C = \mathbf{R}
  \left(
    \mathbf{p} - \left[c_x^{V_i}, c_y^{V_i}, c_z^{V_i}\right]^{\rm T}
  \right)
  \label{eq:canonicalization}
\end{equation}
where $c_x^{V_i}$, $c_y^{V_i}$, $c_z^{V_i}$ represent the center coordinates of the vehicle $V_i$ along the X, Y, and Z axes, respectively.
$\mathbf{R}$ is the rotation matrix used to normalize the 3D point into the canonical feature space.
\begin{equation}
  \mathbf{R} = 
  \begin{bmatrix}
      \cos\theta & \sin\theta & 0 \\
      -\sin\theta \cos\gamma & \cos\theta \cos\gamma & \sin\gamma \\
      \sin\theta \sin\gamma & -\cos\theta \sin\gamma & \cos\gamma
    \end{bmatrix}
\end{equation}
where $\theta \in (-180\degree, 180\degree]$ denotes the yaw angle, indicating the vehicle's heading in the XY-plane relative to the $-y$-axis, while $\gamma \in (-90\degree, 90\degree)$ represents the pitch angle, with positive or negative values indicating upward or downward tilt relative to the XY-plane.
The feature $\mathbf{f}_{\rm V_i}^{(\mathbf{p}^C, t)}$ corresponding to the vehicle $V_i$ at time step $t$ for $\mathbf{p}^C$ is derived as
\begin{equation}
  \mathbf{f}_{V_i}^{(\mathbf{p}^C, t)} = \mathcal{O}({\rm Concat}(\mathbf{f}_{V_i}^t, \mathbf{p}^C))
\end{equation}
where $\mathbf{f}_{V_i}^t \in \mathbb{R}^{d_V}$ is the features extracted from the local scene $(\mathbf{H}_{T_t}^{V_i}, \mathbf{S}_{T_t}^{V_i})$ using the global encoder $\textrm{E}_\textrm{V}$.
\begin{equation}
  \mathbf{f}_{V_i}^t = \textrm{E}_\textrm{V}(\mathbf{H}_{T_t}^{V_i}, \mathbf{S}_{T_t}^{\rm V_i})
  \label{eq:vehicle-local-encoder}
\end{equation}

\noindent \textbf{Volumetric Rendering.}
The volumetric rendering mirrors Building Instance Generator, using a style code $\mathbf{z}$ to represent the variability in vehicle appearances.
The pixel value $C(\mathbf{r})$ is calculated through an integration process as described in Equation~\ref{eq:vol-render-z}.
The camera ray $\mathbf{r}(t)$  is normalized to the canonical feature space following Equation~\ref{eq:canonicalization}.

\noindent \textbf{Loss Function.}
Vehicle Instance Generator is optimized with a hybrid objective that integrates reconstruction and adversarial objectives. Specifically, the training process incorporates an $L_1$ loss, a perceptual loss $\mathcal{P}$, and a GAN loss $\mathcal{G}$ to balance fidelity and realism.
\begin{equation}
  \ell_V = \lambda_V^{\rm L1} \lVert\mathbf{\hat{I}}_{V_i}^t - \mathbf{I}_{V_i}^t\rVert
         + \lambda_V^{\rm P} \mathcal{P}(\mathbf{\hat{I}}_{V_i}^t, \mathbf{I}_{V_i}^t)
         + \lambda_V^{\rm G} \mathcal{G}(\mathbf{\hat{I}}_{V_i}^t, \mathbf{S}_{V_i}^t)
\end{equation}
where $\mathbf{I}_{V_i}^t$ denotes the ground truth image of the vehicle instance $V_i$ at time step $t$, while $\mathbf{S}_{V_i}^t$ is the corresponding perspective-view semantic map, generated in a manner similar to $\mathbf{S}_G$. 
The weights of the three losses are represented as $\lambda_V^{\rm L1}$, $\lambda_V^{\rm P}$, and $\lambda_V^{\rm G}$. 
Note that $\ell_V$ is applied exclusively to pixels with semantic labels belonging to the vehicle instance.

\subsection{Compositor}
\label{sec:compositor}
As there are no ground truth images available for the outputs generated by City Background Generator, Building Instance Generator, and Vehicle Instance Generator, training neural networks to combine these images becomes challenging.
Consequently, Compositor merges the generated images and their corresponding masks into one unified image.
\begin{equation}
  \mathbf{I}_C^t = \mathbf{\hat{I}}_G\mathbf{M}_G
                 + \sum_{i=1}^{n_B} {\mathbf{\hat{I}}_{B_i}\mathbf{M}_{B_i}}
                 + \sum_{i=1}^{n_V} {\mathbf{\hat{I}}_{V_i}^t\mathbf{M}_{V_i}^t}
\end{equation}

\section{Datasets}

\begin{figure}[!t]
  \includegraphics[width=\linewidth]{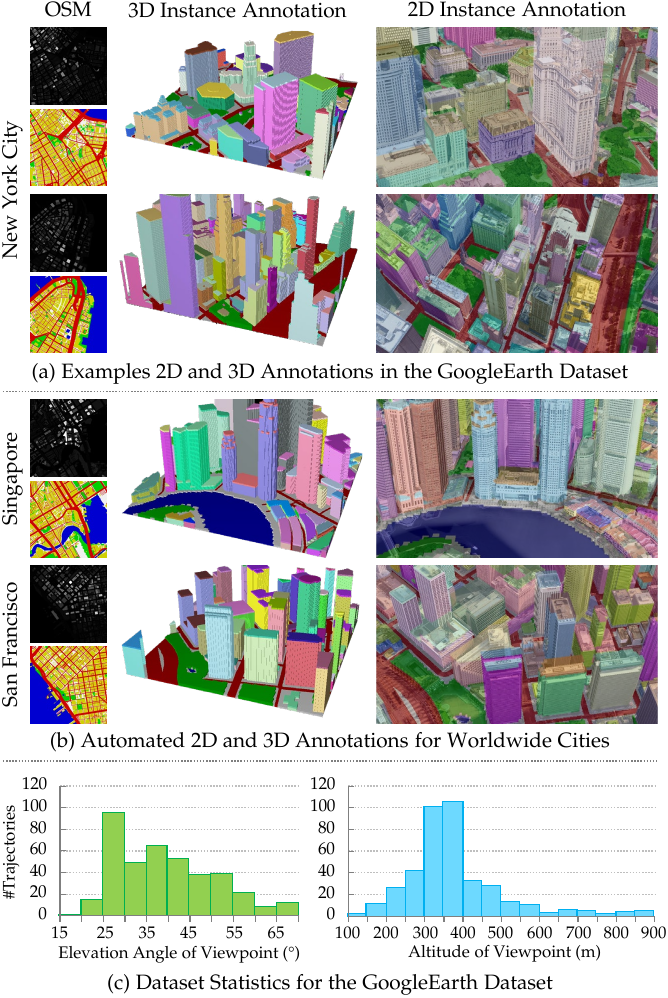}
  \caption{\textbf{Overview of the OSM and GoogleEarth Datasets.} (a) Examples of the 2D and 3D annotations in the GoogleEarth dataset, which can be automatically generated using the OSM dataset. (b) The automatic annotation pipeline can be readily adapted for worldwide cities. (c) The dataset statistics highlight the diverse perspectives in the GoogleEarth dataset.}
  \label{fig:google-earth-dataset}
\end{figure}

\begin{figure}[!t]
  \includegraphics[width=\linewidth]{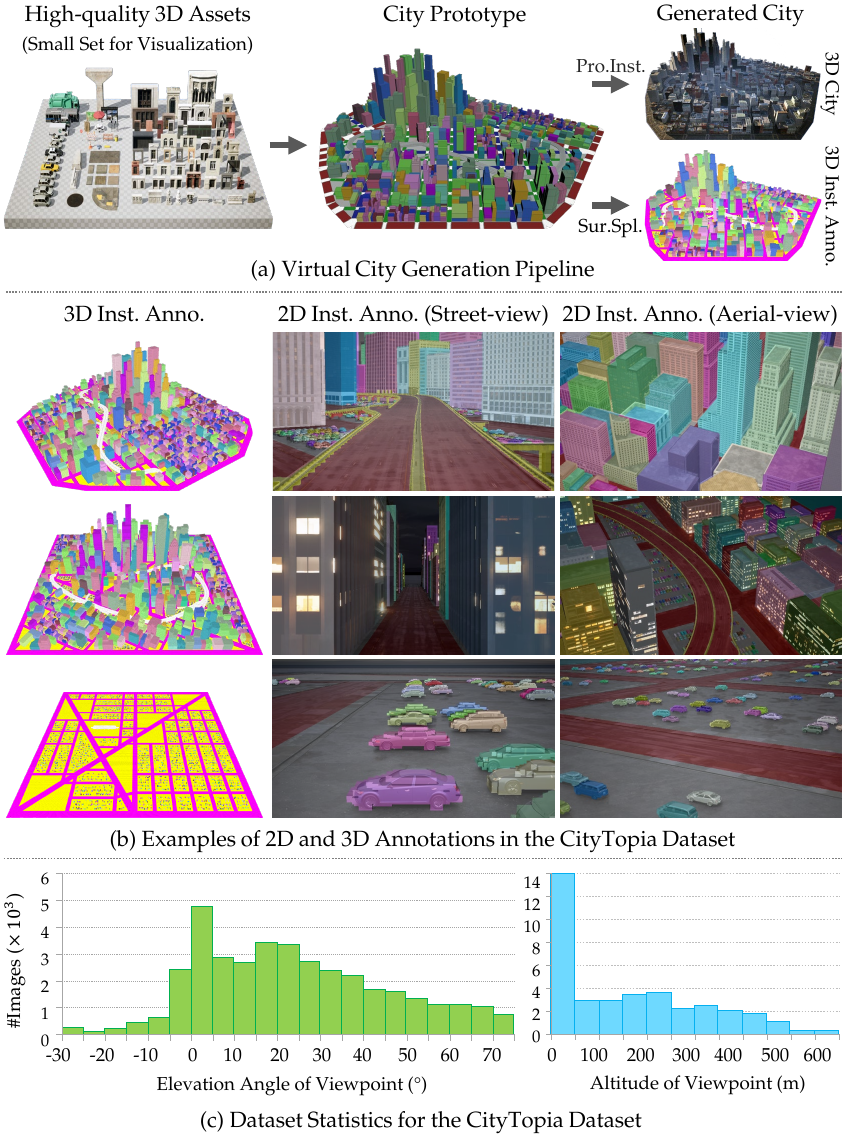}
  \caption{\textbf{Overview of the CityTopia Dataset.} (a) The virtual city generation pipeline. ``Pro.Inst.'', ``Sur.Spl'', and ``3D Inst. Anno.'' denote ``Prototype Instantiation'', ``Surface Sampling'', and ``3D Instance Annotation'', respectively. (b) Examples of 2D and 3D annotations in the CityTopia dataset are shown from both daytime and nighttime street-view and aerial-view perspectives, automatically generated during virtual city generation. (c) The dataset statistics highlight the diverse perspectives in both street and aerial views.}
  \label{fig:citytopia-dataset}
\end{figure}

\subsection{OSM Dataset}

The OSM dataset, collected from OpenStreetMap~\cite{HAOZHE:link/OpenStreetMap}, includes rasterized semantic maps and height fields for 80 cities across the globe, covering more than 6,000 km\textsuperscript{2}. 
In the rasterization step, vector data is transformed into images by converting longitude and latitude coordinates into the EPSG:3857 coordinate system at zoom level 18, which gives a resolution of approximately 0.597 meters per pixel. 
As shown in Fig.~\ref{fig:google-earth-dataset}, The segmentation maps use different colors to indicate various elements: red for roads, yellow for buildings, green for urban greenery, cyan for construction areas, and blue for water bodies. 
The height fields mainly capture building elevations, based on OpenStreetMap data. 
The heights for roads are set to 4, water bodies at 0, and urban greenery is assigned random heights, generated using Perlin noise~\cite{DBLP:conf/siggraph/Perlin85} within a range of 8 to 16 meters.

\subsection{GoogleEarth Dataset}

\modelname ~generates each building instance in the city separately to handle the diversity of buildings, which requires dense 3D instance annotations.
As shown in Table~\ref{tab:comp-dataset}, no existing dataset provides both dense 3D semantic and instance annotations.
To address this, we automatically generate dense 3D semantic and building instance annotations for the GoogleEarth dataset by geographically aligning Google Earth and OpenStreetMap using latitude and longitude.

\noindent \textbf{Image Collection.}
The GoogleEarth dataset, collected from Google Earth Studio~\cite{HAOZHE:link/GoogleEarth}, includes 400 orbit trajectories over the New York City, totaling 24,000 images at a 960x540 resolution. 
As shown in Fig.~\ref{fig:google-earth-dataset}\hyperref[fig:google-earth-dataset]{c}, orbit radii range from 125 to 813 meters, with altitudes from 112 to 884 meters. 
Google Earth Studio also provides camera intrinsic and extrinsic parameters for each image.

\noindent \textbf{2D and 3D Annotation.}
The 3D annotations can be generated by: 
\textbf{1)} performing connected components detection on the OSM semantic map to create the instance map for buildings, while keeping the labels for background stuff unchanged, and 
\textbf{2)} generating 3D volumes by extruding the pixels in the instance map based on height values from the OSM dataset.
The dense 3D annotations can be used to create 2D annotations by projecting the 3D volumes onto images, leveraging the camera parameters from Google Earth Studio.
Fig.~\ref{fig:google-earth-dataset}\hyperref[fig:google-earth-dataset]{a} shows the 2D and 3D instance annotations in the GoogleEarth dataset, highlighting the efficiency of automated data annotation.  
Fig.~\ref{fig:google-earth-dataset}\hyperref[fig:google-earth-dataset]{b} shows how the automated annotation pipeline can be applied to cities worldwide.

\begin{table*}
  \setlength\tabcolsep{4pt}
  \setlength\extrarowheight{2pt}
  \caption{\textbf{Quantitative Comparison.} The best values are highlighted in bold. Note that InfiniCity is not included in this comparison as it is not open-sourced.}
  \label{tab:comp-citygen}
  \begin{tabularx}{\linewidth}{cYYYYY|YYYYY}
    \toprule
      \multirow{2}{*}{Methods} &
      \multicolumn{5}{c|}{GoogleEarth} &
      \multicolumn{5}{c}{CityTopia} \\
      \cline{2-6} \cline{7-11}
            & FID \dwar & KID \dwar & VBench \upar & DE \dwar & CE \dwar & 
              FID \dwar & KID \dwar & VBench \upar & DE \dwar & CE \dwar \\
    \midrule
    SGAM~\cite{DBLP:conf/nips/ShenMW22} &  
              277.6     & 0.358      & 0.691        & 0.575      & 239.2      &  
              330.1     & 0.284      & 0.690        & 0.571      & 233.5 \\
    PersistentNature~\cite{DBLP:conf/cvpr/Chai0LIS23} &  
              123.8     & 0.109      & 0.706        & 0.326      & 86.37      &  
              235.3     & 0.215      & 0.713        & 0.428      & 127.3 \\
    SceneDreamer~\cite{DBLP:journals/pami/ChenWL23} &  
              232.2     & 0.204      & 0.781        & 0.153      & 0.186      &  
              195.1     & 0.126      & 0.708        & 0.185      & 0.162 \\
    DreamScene4D~\cite{DBLP:conf/nips/ChuKF24} & 
              -         & -          & -            & -          & -          &  
              288.2     & 0.136      & 0.715        & 0.199      & 0.146 \\
    DimensionX~\cite{DBLP:preprint/arxiv/2411-04928} & 
            206.9       & 0.182      & 0.805        & -          & -          &  
            171.4       & 0.070      & 0.815        & -          & - \\
    \modelname~(Ours) &  
             \bf{96.83} & \bf{0.096} & \bf{0.834}   & \bf{0.138} & \bf{0.060} &  
             \bf{88.48} & \bf{0.049} & \bf{0.825}   & \bf{0.150} & \bf{0.063} \\
    \bottomrule
  \end{tabularx}
\end{table*}


\subsection{CityTopia Dataset}

The GoogleEarth dataset provides images with dense 3D semantic and instance annotations but faces three challenges: 
\textbf{1)} it lacks street-view images due to suboptimal 3D reconstructions near ground level in Google Earth Studio~\cite{HAOZHE:link/GoogleEarth}; 
\textbf{2)} its annotations, sourced from OpenStreetMap~\cite{HAOZHE:link/OpenStreetMap}, have some imprecision due to differing data sources; and 
\textbf{3)} elevated structures like highways remain unannotated due to missing height data in OpenStreetMap.
To address these challenges, we construct the CityTopia dataset, featuring precise 3D dense annotations on high-fidelity day and night images from both street and aerial views.
As shown in Table~\ref{tab:comp-dataset}, it is the largest dataset to date, offering unparalleled scene diversity and detailed annotations for urban cities.

\noindent \textbf{Virtual City Generation.}
To build the CityTopia dataset, we design 11 virtual cities in Houdini and Unreal Engine, generating 3D annotations and realistic images with controlled lighting conditions.
As illustrated in Fig.~\ref{fig:citytopia-dataset}\hyperref[fig:citytopia-dataset]{a}, we use a diverse, high-quality set of approximately 5,000 3D assets from the CitySample project~\cite{HAOZHE:link/CitySample} to procedurally generate a city prototype in Houdini\footnote{\url{https://www.sidefx.com}}. 
This city prototype stores the 6D poses of all 3D assets within the city. 
Through surface sampling, we can assign each 3D point a semantic and instance label, and by instantiating the city prototype in Unreal Engine\footnote{\url{https://www.unrealengine.com}}, we produce a fully generated virtual city.

\noindent \textbf{Image Collection.}
Once the virtual city is instantiated in Unreal Engine, camera trajectories are set to generate 3,000 images for cities with buildings and 7,500 for a vehicle-only city.
Daytime and nighttime scenes are rendered for each trajectory, with sunlight removed to help the network more easily learn lighting consistency during the generation process.
To avoid Moir\'e effects, each image is sampled 8x spatially and 32x temporally during rendering.
As shown in Fig.~\ref{fig:citytopia-dataset}\hyperref[fig:citytopia-dataset]{c}, the CityTopia dataset provides a wider range of viewpoints, shown by its broader elevation angles compared to the GoogleEarth dataset, as well as more street-level perspectives, evidenced by the large number of images taken at near-zero altitude.

\noindent \textbf{2D and 3D Annotation.}
Since the precise 3D annotations are natively generated from the virtual city pipeline, once the camera poses are set in Unreal Engine, 2D annotations are produced by projecting the 3D annotations using the given camera poses.
Fig.~\ref{fig:citytopia-dataset}\hyperref[fig:citytopia-dataset]{b} highlights the perfect alignment of 2D and 3D instance annotations with both street-view and aerial-view images in the CityTopia dataset. 
The last row features a vehicle-only scene, enhancing vehicle generation learning.
The accurate vehicle annotations demonstrate the effectiveness of the pipeline, which can be scaled by adding more 3D assets.

\begin{figure*}[!t]
    \includegraphics[width=\linewidth]{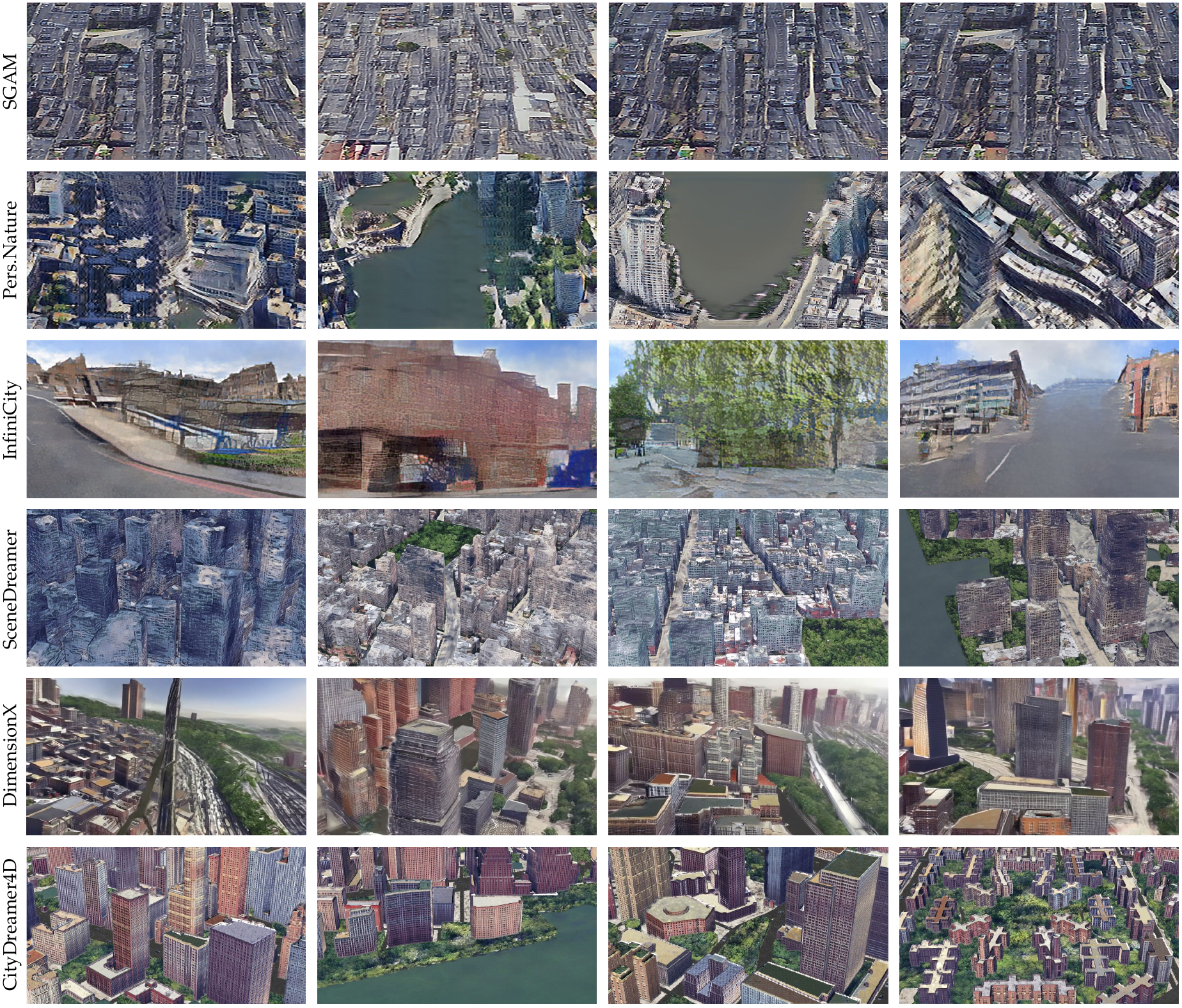}
    \caption{\textbf{Qualitative Comparison on Google Earth.} For SceneDreamer~\cite{DBLP:journals/pami/ChenWL23} and \modelname, vehicles are generated using models trained on CityTopia due to the lack of semantic annotations for vehicles in Google Earth. For DimensionX~\cite{DBLP:preprint/arxiv/2411-04928}, the initial frame is provided by \modelname. The visual results of InfiniCity~\cite{DBLP:conf/iccv/LinLMCS0T23}, provided by the authors, have been zoomed in for better viewing. ``Pers.Nature'' stands for ``PersistentNature''~\cite{DBLP:conf/cvpr/Chai0LIS23}.}
    \label{fig:comp-google-earth}
\end{figure*}

\begin{figure*}[!t]
    \includegraphics[width=\linewidth]{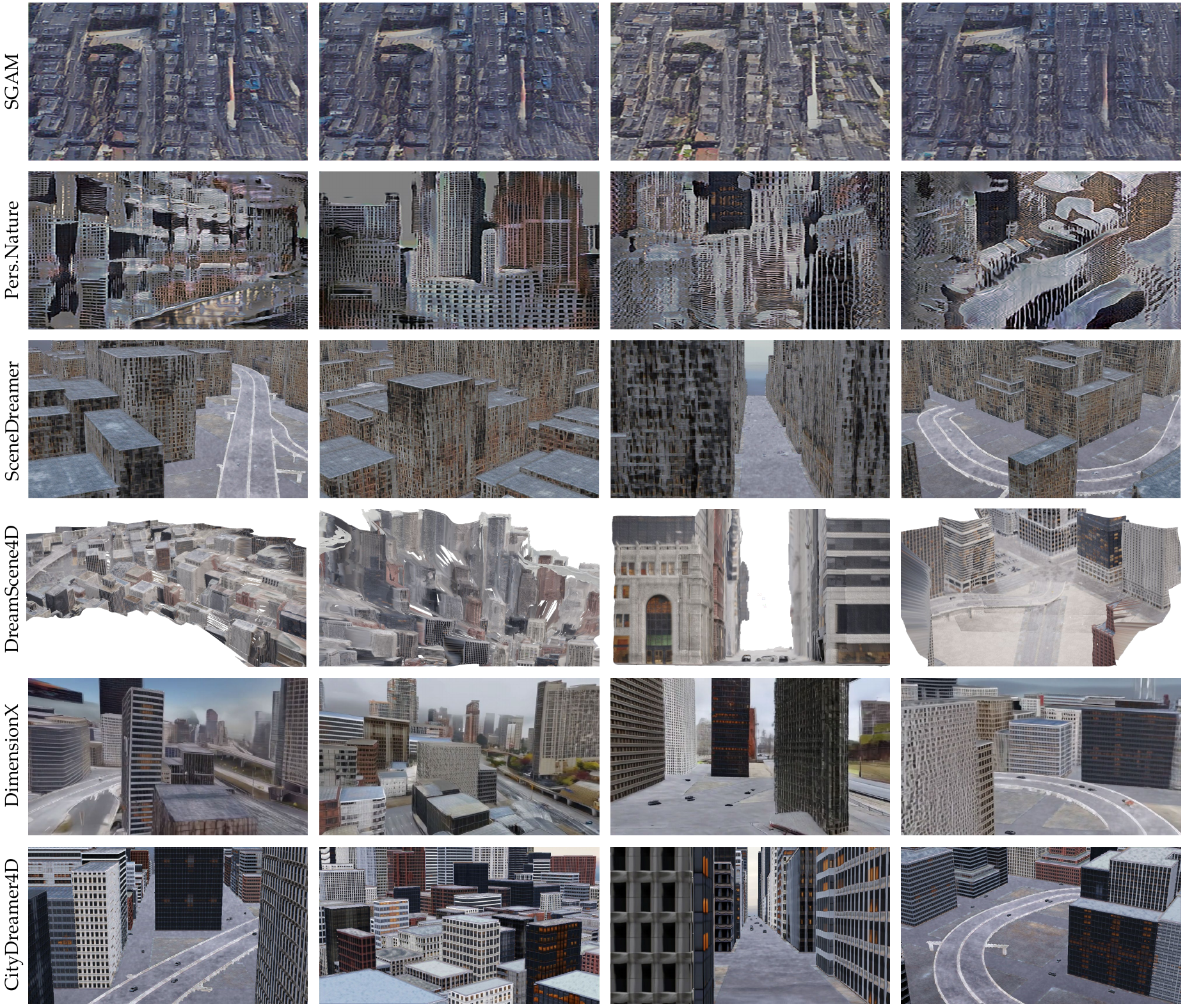}
    \caption{\textbf{Qualitative Comparison on CityTopia.} The initial frame for DimensionX and the input frames for DreamScene4D are chosen from the dataset. ``Pers.Nature'' refers to ``PersistentNature''~\cite{DBLP:conf/cvpr/Chai0LIS23}.}
    \label{fig:comp-citytopia}
\end{figure*}

\section{Experiments}

\subsection{Evaluation Protocols}

We evaluate our method by generating 1,024 unique city layouts, each with 20 variations created by randomly sampling the style code $\mathbf{z}$. 
For each variation, images are rendered at a resolution of $960 \times 540$ pixels using randomized camera trajectories. 
Frames from these renderings are randomly selected for evaluation, depending on the specific metrics used. 
The evaluation metrics are as follows.

\noindent \textbf{FID and KID}.
Fr\'echet Inception Distance (FID)~\cite{DBLP:conf/nips/HeuselRUNH17} and Kernel Inception Distance (KID)~\cite{DBLP:conf/iclr/BinkowskiSAG18} measure image quality. 
FID and KID are calculated between 15,000 generated frames and 15,000 randomly sampled images from datasets.

\noindent \textbf{VBench}.
VBench~\cite{DBLP:conf/cvpr/HuangHYZS0Z0JCW24}  provides a comprehensive evaluation of video generative models, considering dimensions such as background consistency, motion smoothness, dynamic degree, aesthetic quality, and imaging quality. 
The VBench score is computed from 150 videos, each consisting of 100 frames rendered at 16 FPS.

\noindent \textbf{Depth Error (DE)}.
To assess 3D geometry, DE is evaluated following EG3D~\cite{DBLP:conf/cvpr/ChanLCNPMGGTKKW22}.
A pretrained model~\cite{DBLP:journals/pami/RanftlLHSK22} generates pseudo ground truth depth maps by accumulating density $\sigma$. 
DE is calculated as the L2 distance between the normalized depth maps, evaluated on 100 frames per method.

\noindent \textbf{Camera Error (CE)}.
CE measures multi-view consistency, following SceneDreamer~\cite{DBLP:journals/pami/ChenWL23}. 
CE is computed on a static 3D scene by comparing the inferred camera trajectory with the one estimated by COLMAP~\cite{DBLP:conf/cvpr/SchonbergerF16}. 
This metric is calculated on 600 frames rendered from an orbit trajectory and is defined as the scale-invariant normalized L2 distance between the generated and reconstructed camera poses.

\subsection{Implementation Details}

\noindent \textbf{Hyperparameters}

\noindent \textit{Unbounded Layout Generator.}
The codebook size $d_K$ is set to $512$, with each code having a dimension $d_C$ of $512$.
Height field and semantic map patches are cropped to 512$\times$512 and compressed by a factor of $16$.
The loss weights are $\lambda_{\rm R} = 10$, $\lambda_{\rm S} = 10$, and $\lambda_{\rm E} = 1$.

\noindent \textit{City Background Generator.}
For the GoogleEarth dataset, the local window resolutions are set to $N_G^H = 1536$, $N_G^W = 1536$, and $N_G^D = 640$. 
For the CityTopia dataset, they are set to $N_G^H = 3072$, $N_G^W = 3072$, and $N_G^D = 2560$.
The dimension of scene-level features $d_G$ is $2$. 
For the generative hash grid, $N_H^L = 16$, $N_E = 2^{19}$, and $N_G^C = 8$.
The prime numbers used in Equation~\ref{eq:hashgrid} are $\pi^1 = 1$, $\pi^2 = 2654435761$, $\pi^3 = 805459861$, $\pi^4 = 3674653429$, and $\pi^5 = 2097192037$.
The loss function weights are set to $\lambda_G^{\rm L1} = 10$, $\lambda_G^{\rm P} = 10$, and $\lambda_G^{\rm G} = 0.5$.

\noindent \textit{Building Instance Generator.}
For the GoogleEarth dataset, the local window resolutions are set to $N_B^H = 672$, $N_B^W = 672$, and $N_B^D = 640$. 
For the CityTopia dataset, these values are $N_B^H = 768$, $N_B^W = 768$, and $N_B^D = 2560$.
The pixel-level features have 63 channels ($N_B^C = 63$), and the dimension $N_P^L$ is set to $10$.

\noindent \textit{Vehicle Instance Generator.}
The dimension of scene-level features $d_V$ is $2$. 
The local window resolutions are set to $N_V^H = 32$, $N_V^W = 32$, and $N_V^D = 32$. 
The loss function weights are assigned as $\lambda_V^{\rm L1} = 10$, $\lambda_V^{\rm P} = 10$, and $\lambda_V^{\rm G} = 0.5$.

\noindent \textbf{Training Details}

\noindent \textit{Unbounded Layout Generator.}
The VQVAE model is trained over 1,250,000 iterations using a batch size of 16, an Adam optimizer with $\beta = (0.5, 0.9)$, and a learning rate of $7.2\times10^{-5}$. 
The autoregressive transformer is trained for 250,000 iterations with a batch size of 80, an Adam optimizer with $\beta = (0.9, 0.999)$, and a learning rate of $2\times10^{-4}$.

\noindent \textit{Stuff and Instance Generators.}
The City Background Generator, Building Instance Generator, and Vehicle Instance Generator are trained with an Adam optimizer, using $\beta = (0, 0.999)$ and a learning rate of $10^{-4}$. 
The discriminators use the same optimizer settings with a learning rate of $10^{-5}$. 
Training runs for 298,500 iterations with a batch size of $8$, and images are randomly cropped to 192$\times$192 resolution.

\subsection{Main Results}

\noindent \textbf{Comparison Methods.}
We compare \modelname ~against several state-of-the-art methods, including SGAM~\cite{DBLP:conf/nips/ShenMW22}, PersistentNature~\cite{DBLP:conf/cvpr/Chai0LIS23}, SceneDreamer~\cite{DBLP:journals/pami/ChenWL23}, and InfiniCity~\cite{DBLP:conf/iccv/LinLMCS0T23}. 
Since no method exists for 4D scene generation, we use DreamScene4D~\cite{DBLP:conf/nips/ChuKF24} for 4D novel view synthesis and DimensionX~\cite{DBLP:preprint/arxiv/2411-04928} for 4D video generation as competitive baselines.
To ensure a fair comparison, all methods, except for InfiniCity and DimensionX, are retrained using their released code on the GoogleEarth and CityTopia datasets. 
Since SceneDreamer cannot generate city layouts or traffic scenarios, their inputs are supplied by Unbounded Layout Generator and Traffic Scenario Generator. 
Additionally, because the GoogleEarth dataset lacks annotations for dynamic objects, vehicles are generated using models trained on the CityTopia dataset to support 4D generation.

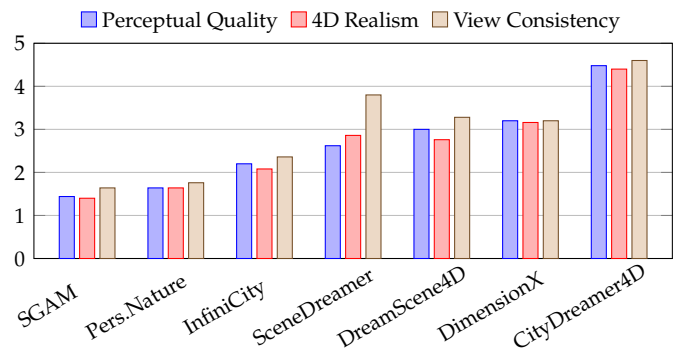
\begin{figure}[!t]
    \begin{tikzpicture}
  \pgfplotstableread[row sep=\\,col sep=&]{
    Method        & PQ   & DR   & VC \\
    SGAM          & 1.44 & 1.40 & 1.64  \\
    Pers.Nature   & 1.64 & 1.64 & 1.76  \\
    InfiniCity    & 2.20 & 2.08 & 2.36  \\
    SceneDreamer  & 2.62 & 2.86 & 3.80  \\
    DreamScene4D  & 3.00 & 2.76 & 3.28  \\
    DimensionX    & 3.20 & 3.16 & 3.20  \\
    \modelname    & 4.48 & 4.40 & 4.60  \\
  }\userStudyData

  \begin{axis}[
    ybar,
    width             = \linewidth,
    height            = .5\linewidth,
    ymajorgrids       = true,
    symbolic x coords = {
      SGAM,
      Pers.Nature,
      InfiniCity,
      SceneDreamer,
      DreamScene4D,
      DimensionX,
      \modelname
    },
    ymin              = 0,
    ymax              = 5,
    x                 = 7.5 ex,
    bar width         = 2 mm,
    y label style     = {at = {(-0.03, 0.5)}},
    x tick style      = {draw = none},
    xticklabel style  = {
      align           = right,
      font            = \fontsize{8}{8}\selectfont,
      rotate          = 30,
      anchor          = base,
      xshift          = -7 mm,
      yshift          = -2.5 mm,
    },
    ytick             = {0, 1, 2, 3, 4, 5},
    yticklabels       = {0, 1, 2, 3, 4, 5},
    yticklabel style  = {
      align           = center,
      font            = \fontsize{8}{8}\selectfont,
    },
    legend image code/.code={
      \draw [#1] (0 mm, -0.8 mm) rectangle (2 mm, 1.2 mm);
    },
    legend style      = {
      at              = {(0.5, 1.2)},
      anchor          = north,
      draw            = none,
      font            = \fontsize{8}{8}\selectfont,
      legend columns  = -1,
      /tikz/every even column/.append style={column sep = 1 mm}
    }]
    \addplot table[x = Method,y = PQ]{\userStudyData};
    \addplot table[x = Method,y = DR]{\userStudyData};
    \addplot table[x = Method,y = VC]{\userStudyData};

    \addlegendentry{Perceptual Quality};
    \addlegendentry{4D Realism};
    \addlegendentry{View Consistency};
  \end{axis}
\end{tikzpicture}
    \vspace{-3.5 mm}
    \caption{\textbf{User Study on 4D City Generation.} All scores are in the range of 5, with 5 indicating the best. ``Pers.Nature'' refers to ``PersistentNature''~\cite{DBLP:conf/cvpr/Chai0LIS23}.}
    \label{fig:user-study}
\end{figure}

\begin{figure*}[!t]
    \centering
    \includegraphics[width=\linewidth]{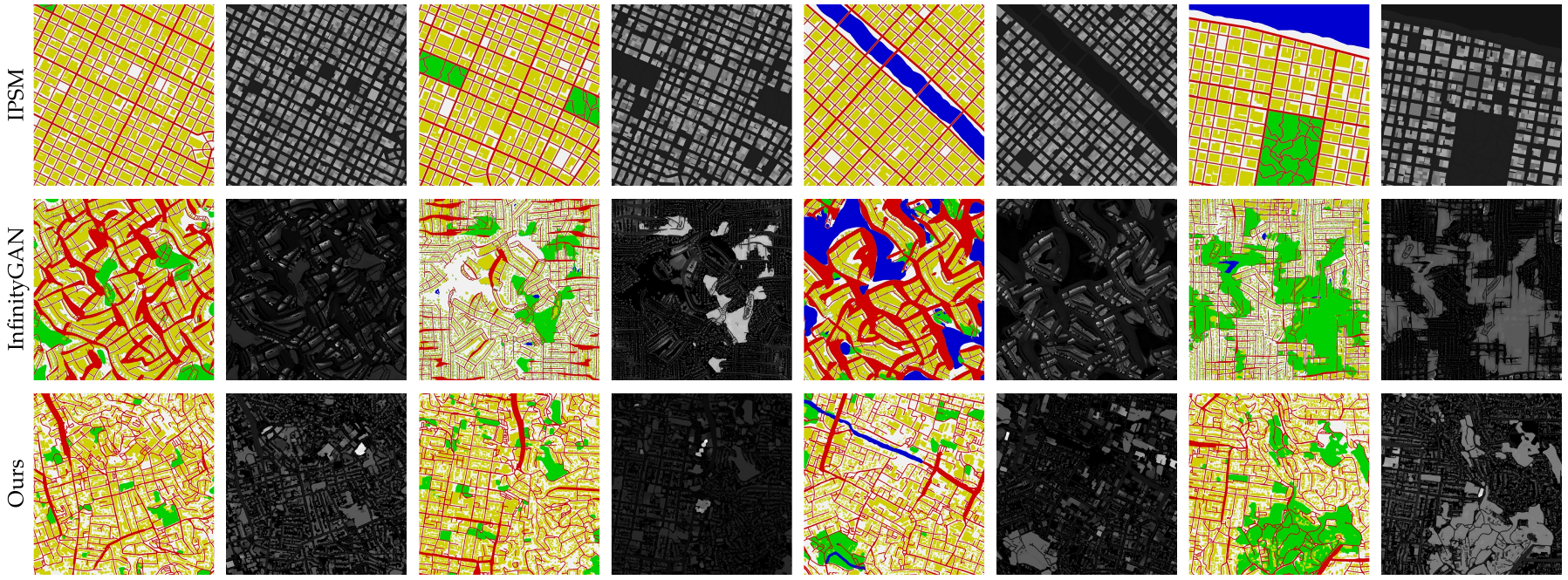}
    \caption{\textbf{Qualitative Comparison of City Layout Generators.} The height map values are normalized to a range of $[0, 1]$ by dividing each value by the maximum value within the map.}
    \label{fig:abs-layout-gen}
\end{figure*}

\noindent \textbf{Qualitative Comparison.}
Fig.~\ref{fig:comp-google-earth} and~\ref{fig:comp-citytopia} present qualitative comparisons with the baseline methods on the GoogleEarth and CityTopia datasets, respectively.
SGAM faces difficulties in generating realistic results and maintaining multi-view consistency due to the inherent challenges of extrapolating views for complex 4D cities.
PersistentNature, which adopts a tri-plane representation, also struggles to produce realistic renderings.
Both InfiniCity and SceneDreamer use BEV maps as their scene representation, but they still experience significant structural distortions in instance-level objects, such as buildings and vehicles, because all instances are assigned the same semantic label.
DreamScene4D cannot directly generate 4D scenes but transforms monocular videos into 4D scenes by decoupling dynamic objects from the background, yet it struggles to reconstruct their 3D shapes.
During the generation of orbit 4D videos, DimensionX exhibited severe distortions and failed to maintain multi-view consistency in the results.
In comparison, the proposed \modelname ~generates more realistic and diverse results compared to all the baselines\footnote{More results can be found on our project page.}.

\noindent \textbf{Quantitative Comparison.}
Table~\ref{tab:comp-citygen} shows the quantitative metrics, where \modelname ~outperforms the baselines in FID, KID, and VBench, highlighting its motion smoothness, dynamic degree, and aesthetic quality. 
Additionally, \modelname ~achieves the lowest DE and CE errors, demonstrating accurate 3D geometry, view consistency, and photorealistic image generation.

\noindent \textbf{User Study.}
To better evaluate the multi-view consistency and quality of unbounded 4D city generation, we perform a user study following CityDreamer's protocol~\cite{DBLP:conf/cvpr/XieCHL24}.
In this survey, 25 volunteers rate each generated city on three aspects: 1) perceptual quality, 2) 4D realism, and 3) view consistency. 
Ratings are on a scale of 1 to 5, with 5 being the highest. 
As shown in Fig.~\ref{fig:user-study}, the proposed \modelname ~outperforms the baselines by a significant margin.

\subsection{Ablation Studies}

\begin{table}[!t]
  \centering
  \caption{\textbf{Quantitative Comparison of Unbounded Layout Generator (ULG).} The best values are highlighted in bold. The generated images are centrally cropped to a size of 4096$\times$4096.}
  \label{tab:abs-layout-gen}
  \begin{tabularx}{\linewidth}{c|YY}
     \toprule
     Methods     & FID \dwar   & KID \dwar \\ 
     \midrule
     IPSM~\cite{DBLP:journals/tog/ChenEWMZ08}
                 & 321.47      & 0.502      \\ 
     InfinityGAN~\cite{DBLP:conf/iclr/LinLCT022}
                 & 183.14      & 0.288      \\ 
     ULG (Ours)  & \bf{124.45} & \bf{0.123} \\ 
     \bottomrule
  \end{tabularx}
\end{table}


\noindent \textbf{Effectiveness of Unbounded Layout Generator.}
Unbounded Layout Generator (ULG) is essential for producing ``unbounded'' city layouts. 
To demonstrate the effectiveness of ULG, we evaluate its performance against InfinityGAN~\cite{DBLP:conf/iclr/LinLCT022}, which is utilized in InfiniCity, alongside the rule-based city layout generation technique, IPSM~\cite{DBLP:journals/tog/ChenEWMZ08}.
Following InfiniCity~\cite{DBLP:conf/iccv/LinLMCS0T23}, we use FID and KID to quantitatively 
evaluate the quality of the generated layouts.
As illustrated in Table~\ref{tab:abs-layout-gen}, ULG achieves the best results in terms of all metrics compared to IPSM and InfinityGAN.
The qualitative results shown in Fig.~\ref{fig:abs-layout-gen} also demonstrate the high quality and diversity of the proposed method.

\begin{table}[!t]
  \setlength\tabcolsep{4pt}
  \setlength\extrarowheight{2pt}
  \caption{\textbf{Quantitative Comparison of Building Instance Generator Variants.} The best values are highlighted in bold. Note that ``Inst.'' and ``Pos.Enc.'' refer to ``Instance Labels'' and ``Positional Encoding'', while ``G'' and ``L'' denote ``Global Encoder'' and ``Local Encoder'', respectively.}
  \begin{tabularx}{\linewidth}{cc|cc|cc|cccc}
     \toprule
     \multirow{2}{*}{BIG}         & \multirow{2}{*}{Inst.}    & 
     \multicolumn{2}{c|}{Encoder} & \multicolumn{2}{c|}{Pos.Enc.} & 
     \multicolumn{4}{c}{Evaluation Metrics} \\
     \cline{3-4} \cline{5-6} \cline{7-10}
     & & 
     G           & L          & Hash       & SinCos & 
     FID \dwar   & KID \dwar  & DE \dwar   & CE \dwar \\
     \midrule
     \xmark      & \xmark     & -          & -           & -           & -      &
     195.1       & 0.126      & 0.185      & 0.162 \\
     \cmark      & \xmark     & \xmark     & \cmark      & \xmark      & \cmark &
     167.8       & 0.094      & 0.157      & 0.087 \\
     \midrule
     \cmark      & \cmark     & \cmark     & \xmark      & \cmark      & \xmark &
     196.8       & 0.124      & 0.165      & 0.159 \\
     \cmark      & \cmark     & \cmark     & \xmark      & \xmark      & \cmark & 
     197.9       & 0.132      & 0.162      & 0.152\\
     \cmark      & \cmark     & \xmark     & \cmark      & \cmark      & \xmark & 
     182.3       & 0.111      & 0.155      & 0.092 \\
     \cmark      & \cmark     & \xmark     & \cmark      & \xmark      & \cmark &
     \bf{88.48}  & \bf{0.049} & \bf{0.150} & \bf{0.063} \\
     \bottomrule
  \end{tabularx}
  \label{tab:abs-bldg-ins-gen}
\end{table}
\begin{figure}[!t]
    \centering
    \includegraphics[width=\linewidth]{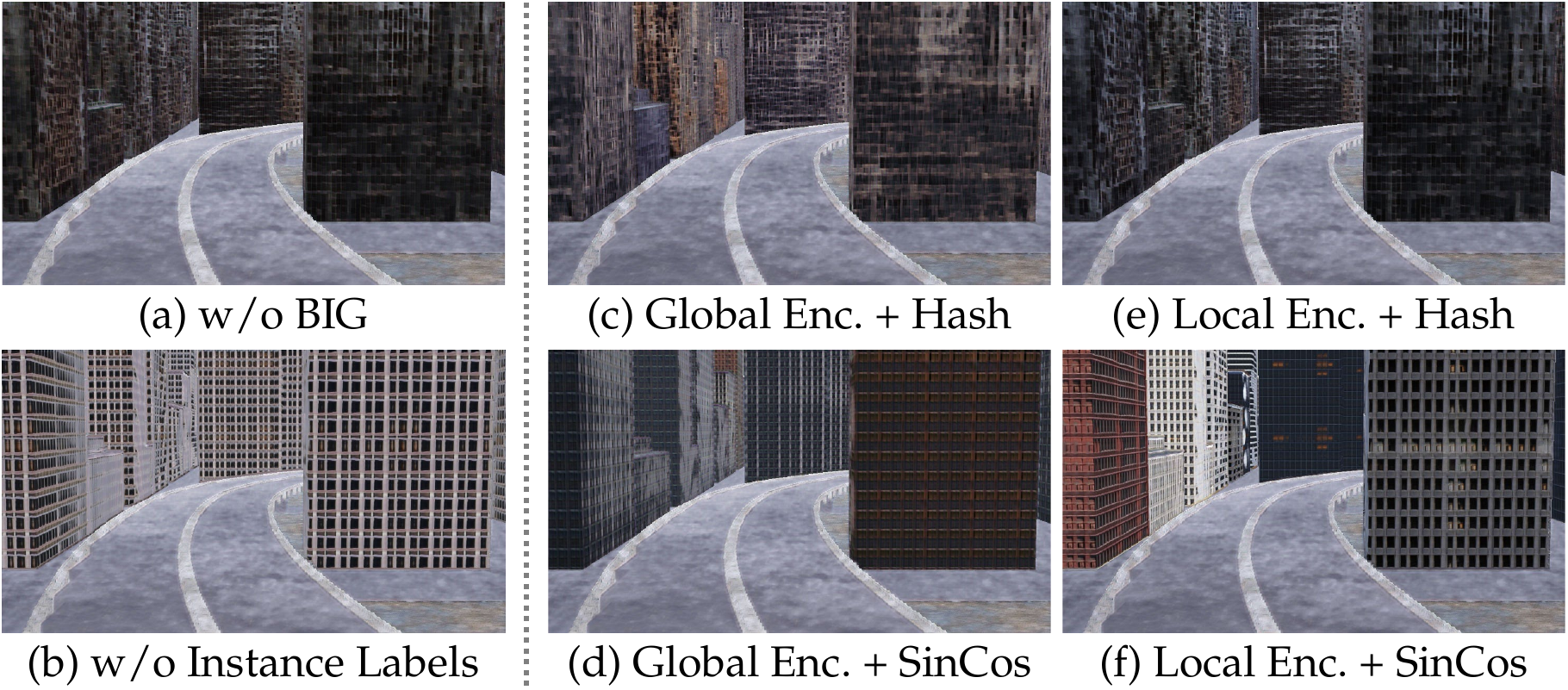}
    \caption{\textbf{Qualitative Comparison of Building Instance Generator (BIG) Variants.} (a) and (b) illustrate the effects of removing BIG and instance labels, respectively. (c)–(f) present the results of various scene parameterizations. Note that ``Enc.'' is an abbreviation for ``Encoder''.}
    \label{fig:abs-bldg-ins-gen}
\end{figure}

\noindent \textbf{Effectiveness of Building Instance Generator.}
We highlight the essential role of Building Instance Generator (BIG) in achieving successful unbounded 4D city generation.
To validate its effectiveness, we perform an ablation study for BIG.
We first compare BIG with two alternative designs: 
(1) Removing BIG from \modelname, effectively reverting the model to SceneDreamer, and 
(2) Generating all buildings simultaneously using BIG without incorporating instance labels.
As shown in the first two rows of Table~\ref{tab:abs-bldg-ins-gen} and Fig.~\ref{fig:abs-bldg-ins-gen}\hyperref[fig:abs-bldg-ins-gen]{a}-\hyperref[fig:abs-bldg-ins-gen]{b}, both alternative designs result in significant degradation in generation quality, underscoring the importance of BIG and instance labels.
Scene parameterization directly impacts the quality of 4D city generation.
BIG uses vanilla SinCos positional encoding with pixel-wise features from the local encoder.
To demonstrate the effectiveness of the scene parameterization in BIG, we compare BIG with other alternative scene parameterization designs.
%
\begin{table}[!t]
  \setlength\tabcolsep{4pt}
  \setlength\extrarowheight{2pt}
  \caption{\textbf{Quantitative Comparison of Vehicle Instance Generator Variants.} All metrics are computed on the vehicle-only city from the CityTopia dataset. The best values are highlighted in bold. Note that ``Can.'' and ``Pos.Enc.'' refer to ``Canonicalization'' and ``Positional Encoding'', while ``G'' and ``L'' denote ``Global Encoder'' and ``Local Encoder'', respectively.}
  \begin{tabularx}{\linewidth}{YY|cc|cc|cccc}
     \toprule
     \multirow{2}{*}{VIG}         & \multirow{2}{*}{Can.}    & 
     \multicolumn{2}{c|}{Encoder} & \multicolumn{2}{c|}{Pos.Enc.} & 
     \multicolumn{4}{c}{Evaluation Metrics} \\
     \cline{3-4} \cline{5-6} \cline{7-10}
     & & 
     G           & L           & Hash        & SinCos & 
     FID \dwar   & KID \dwar   & DE \dwar    & CE \dwar \\
     \midrule
     \xmark      & \xmark      & -           & -           & -           & -      &
     419.3       & 0.576       & 0.364       & 1.276 \\
     \cmark      & \xmark      & \cmark      & \xmark      & \xmark      & \cmark &
     273.4       & 0.530       & 0.289       & 0.966 \\
     \midrule
     \cmark      & \cmark      & \cmark      & \xmark      & \cmark      & \xmark & 
     229.2       & 0.428       & 0.259       & 0.989 \\
     \cmark      & \cmark      & \cmark      & \xmark      & \xmark      & \cmark &
     \bf{142.3}  & \bf{0.276}  & \bf{0.202}  & \bf{0.824} \\
     \cmark      & \cmark      & \xmark      & \cmark      & \cmark      & \xmark & 
     273.4       & 0.521       & 0.265       & 0.997 \\
     \cmark      & \cmark      & \xmark      & \cmark      & \xmark      & \cmark & 
     200.5       & 0.403       & 0.332       & 1.117 \\
     \bottomrule
  \end{tabularx}
  \label{tab:abs-veh-ins-gen}
\end{table}
\begin{figure}[!t]
    \centering
    \includegraphics[width=\linewidth]{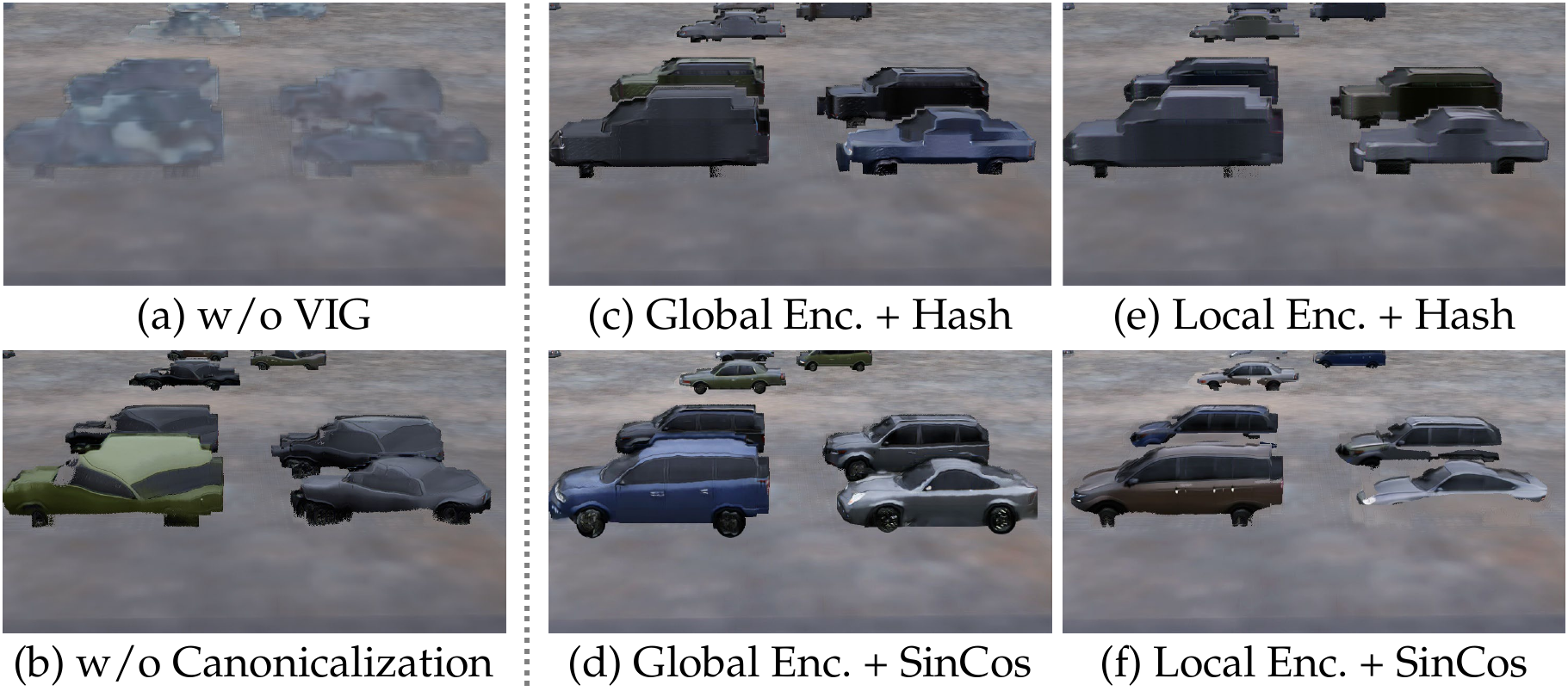}
    \caption{\textbf{Qualitative Comparison of Vehicle Instance Generator (VIG) Variants.} (a) and (b) illustrate the effects of removing VIG and canonicalization, respectively. (c)–(f) present the results of various scene parameterizations. Note that ``Enc.'' is an abbreviation for ``Encoder''.}
    \label{fig:abs-veh-ins-gen}
\end{figure}
%
Scene parameterization plays a critical role in the quality of 4D city generation. 
BIG leverages vanilla SinCos positional encoding combined with pixel-wise features from the local encoder. 
To evaluate the effectiveness of BIG's scene parameterization, we compare it with alternative designs. 
As shown in the last four rows of Table~\ref{tab:abs-bldg-ins-gen} and Fig.~\ref{fig:abs-bldg-ins-gen}\hyperref[fig:abs-bldg-ins-gen]{c}-\hyperref[fig:abs-bldg-ins-gen]{f}, using generative hash grid positional encoding results in distorted building fa\c{c}ades, while Global Encoders with SinCos encoding introduce repetitive fa\c{c}ade patterns. 
These comparisons emphasize the significance of BIG's well-designed parameterization in achieving realistic and varied results.

\noindent \textbf{Effectiveness of Vehicle Instance Generator.}
Vehicle Instance Generator (VIG) plays a critical role in generating vehicles within 4D cities. 
To validate its effectiveness, we conduct an ablation study on VIG.
We compare it with two alternative designs: 
(1) Removing VIG from \modelname ~and treating vehicles as background stuff, allowing City Background Generator to handle their generation, and 
(2) Generating vehicles without canonicalization, meaning they are not produced in a canonical feature space.
As shown in the first two rows of Table~\ref{tab:abs-veh-ins-gen} and Fig.~\ref{fig:abs-veh-ins-gen}\hyperref[fig:abs-veh-ins-gen]{a}-\hyperref[fig:abs-veh-ins-gen]{b}, both alternative designs lead to severe distortions in the generated results, highlighting the importance of VIG and canonicalization.
Scene parameterization is equally critical in VIG. 
To validate this, we compare different scene parameterization designs within VIG.
Currently, VIG uses vanilla SinCos positional encoding combined with global-level features from the global encoder.
In the canonical feature space, combining global-level features with 3D coordinates allows the network to better share features across different vehicles, facilitating better convergence.
As shown in the last row of Table~\ref{tab:abs-veh-ins-gen} and Fig.~\ref{fig:abs-veh-ins-gen}\hyperref[fig:abs-veh-ins-gen]{f}, using a local encoder with SinCos positional encoding, as in BIG, makes learning more challenging, resulting in incomplete vehicle shapes.
Similarly, using generative hash grid in VIG leads to structural distortions by complicating the network's ability to associate texture features with 3D coordinates, as illustrated in Fig.~\ref{fig:abs-veh-ins-gen}\hyperref[fig:abs-veh-ins-gen]{c} and~\ref{fig:abs-veh-ins-gen}\hyperref[fig:abs-veh-ins-gen]{e} as well as the 3rd and 5th rows of Table~\ref{tab:abs-veh-ins-gen}.

\begin{figure}[!t]
    \centering
    \includegraphics[width=\linewidth]{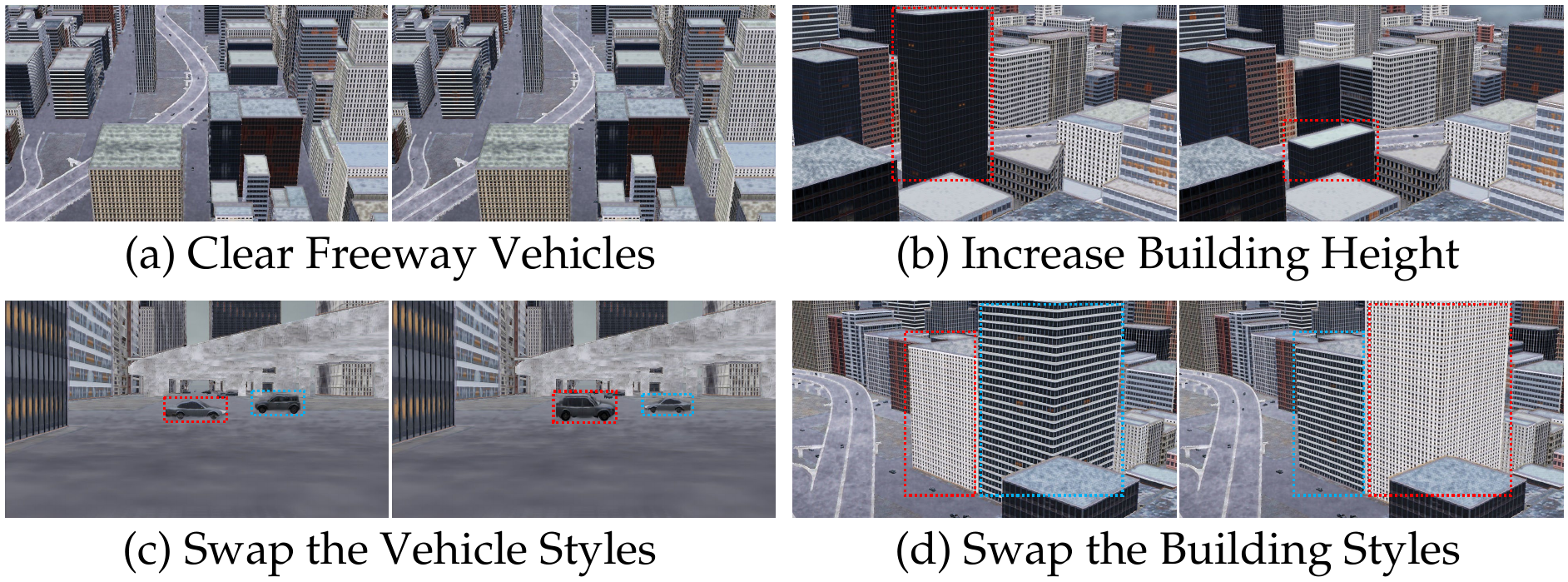}
    \caption{\textbf{Localized Editing on the Generated Cities.} (a) and (c) show vehicle editing results, while (b) and (d) present building editing results.}
    \label{fig:local-editing}
\end{figure}

\begin{figure}[!t]
    \centering
    \includegraphics[width=\linewidth]{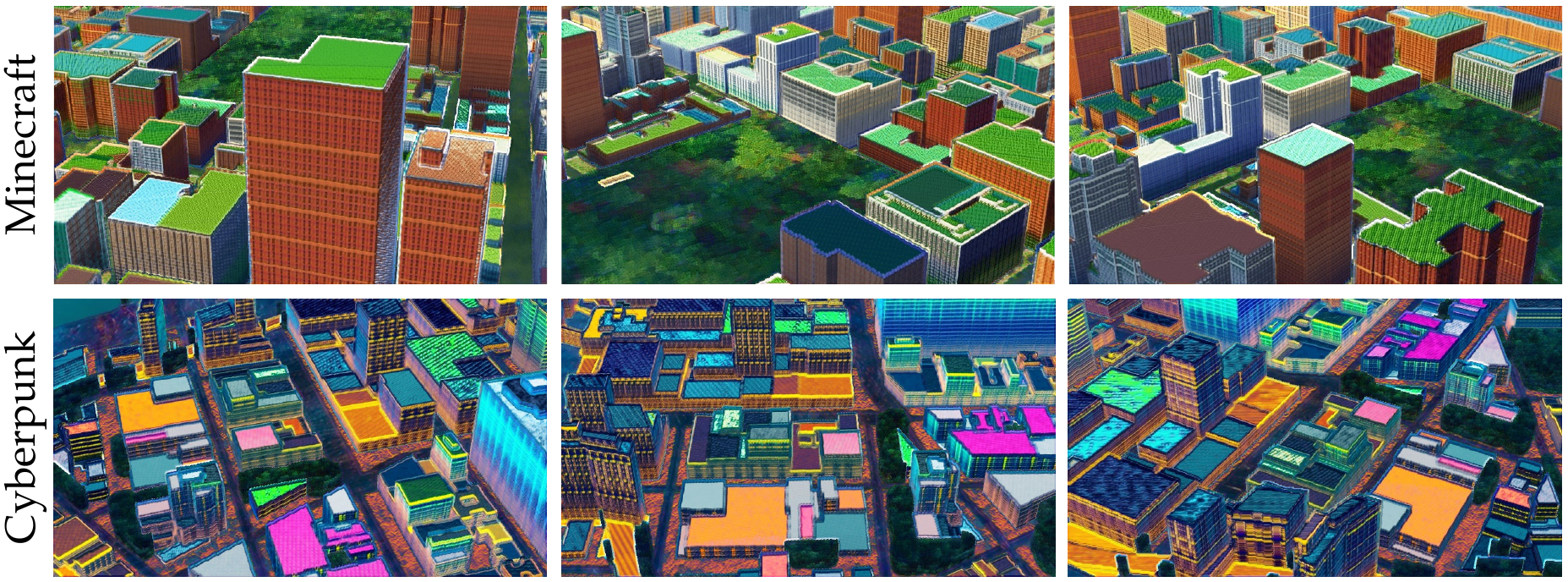}
    \caption{\textbf{Text-driven City Stylization with ControlNet.} The multi-view consistency is preserved in stylized Minecraft and Cyberpunk cities.}
    \label{fig:stylization}
\end{figure}

\subsection{Applications}

\noindent \textbf{Urban Simulator.}
\modelname ~can be a powerful urban simulator, capable of generating realistic 4D urban scenes with dynamic objects and detailed environments. 
Unlike traditional simulators such as CARLA~\cite{DBLP:conf/corl/DosovitskiyRCLK17}, which are limited to predefined, bounded areas, this method supports unbounded urban scenes, creating vast, seamless cityscapes. 
Furthermore, it can generate both street-view and aerial-view perspectives, providing a richer variety of scenarios for applications like autonomous driving, urban planning, and virtual reality.

\noindent \textbf{Localized Editing.}
Benefiting from the compositional architecture, \modelname ~allows for localized editing on building and vehicle instances.
In Fig.~\ref{fig:local-editing}\hyperref[fig:local-editing]{a} and~\ref{fig:local-editing}\hyperref[fig:local-editing]{c}, vehicle positions and styles can be independently modified without affecting other scene elements.
Similarly, as shown in Fig.~\ref{fig:local-editing}\hyperref[fig:local-editing]{b} and~\ref{fig:local-editing}\hyperref[fig:local-editing]{d}, building appearances adapt seamlessly to varying heights while maintaining a consistent style. 
This capability facilitates customized scene refinement in post-production.

\noindent \textbf{City Stylization.}
The generated cities can be seamlessly restyled by leveraging ControlNet~\cite{DBLP:conf/iccv/ZhangRA23}, fine-tuning pretrained models on images created with ControlNet conditioned on HED edges.
Fig.~\ref{fig:stylization} shows examples of city styles such as Minecraft and Cyberpunk. 
These results maintain multiview consistency, enabled by the proposed scene representation and parameterization in \modelname.

\begin{table}
  \caption{\rev{\textbf{Visual Language Navigation (VLN) results in generated 4D cities.} Metrics include: PL (Path Length, scaled by 1/10, in meters), SR (Success Rate, \%), SPL (Success rate weighted by normalized inverse Path Length), and RT (Reset Times). Note that each trajectory is executed five times and is considered a failure if all attempts result in resets.}}
  \label{tab:comp-vln}
  \begin{tabularx}{\linewidth}{lcYYYY}
    \toprule
      Methods & 
      \#Param (B) &  PL\dwar    & SR\upar    & SPL\upar   & RT\dwar \\
    \midrule
      Human (5 participants) &
      -           & 20.73       & 100.0      & 85.87      & 0.00 \\
    \midrule \midrule
      Gemini 2.5 Pro~\cite{DBLP:preprint/arxiv/2312-11805} &
      -           & 9.32        & 12.40      & 4.43       & 0.45 \\
      GPT-4o~\cite{DBLP:preprint/arxiv/2410-21276} &
      -           & 8.97        & \bf{36.00} & \bf{17.32} & \bf{0.11} \\
    \midrule
      SAIL-VL 1.6~\cite{DBLP:conf/acl/DongKYLFR25} & 
      8.33        & 14.56       & 23.40      & 7.63       & 0.28 \\
      Ovis2~\cite{DBLP:preprint/arxiv/2405-20797} & 
      8.94        & 13.96       & 17.00      & 5.01       & 0.35 \\
      Qwen2.5-VL~\cite{DBLP:preprint/arxiv/2502-13923} & 
      8.29        & \bf{5.01}   & 15.00      & 7.01       & 0.37 \\
      Ola~\cite{DBLP:preprint/arxiv/2502-04328} & 
      8.88        & 9.15        & 18.00      & 8.30       & 0.32 \\
      InternVL3~\cite{DBLP:preprint/arxiv/2504-10479} & 
      7.94        & 9.02        & 25.60      & 12.66      & 0.23 \\
    \bottomrule
  \end{tabularx}
\end{table}

\noindent 
\textbf{Visual Language Navigation.}
To evaluate the practicality of our generated 4D urban environments, we conduct experiments on Visual Language Navigation, where an embodied agent navigates based on natural language instructions. 
Specifically, the agent operates within scenes generated by \modelname, pretrained on the CityTopia dataset.
We manually annotate a test set of 100 instruction-trajectory pairs, each guiding the agent to a distinct landmark in the generated scenes. 
Following the protocol in GRUtopia~\cite{DBLP:preprint/arxiv/2407-10943}, the agent receives its current image observation and a language prompt as input to a vision-language model (VLM), which selects one of 12 discrete actions: move forward or diagonally (2/4/6 meters), turn left/right (45°), or stop. 
This process continues iteratively until the model outputs ``stop''.
\rev{We additionally collect a human performance reference from 5 volunteers, who complete the full test set following the same navigation protocol as the VLMs.}
We adopt the zero-shot evaluation setting from GRUtopia, using pre-trained VLMs without task-specific fine-tuning. 
We evaluate the latest state-of-the-art VLMs, including open-source models such as SAIL-VL 1.6~\cite{DBLP:conf/acl/DongKYLFR25}, Ovis2~\cite{DBLP:preprint/arxiv/2405-20797}, Qwen2.5-VL~\cite{DBLP:preprint/arxiv/2502-13923}, Ola~\cite{DBLP:preprint/arxiv/2502-04328}, and InternVL3~\cite{DBLP:preprint/arxiv/2504-10479}, as well as closed-source models like Gemini 2.5 Pro~\cite{DBLP:preprint/arxiv/2312-11805} and GPT-4o~\cite{DBLP:preprint/arxiv/2410-21276}.
Navigation performance is evaluated using four standard metrics: success rate (SR), path length (PL), success rate weighted by normalized inverse path length (SPL), and the number of resets due to occlusions (RT).
As shown in Table~\ref{tab:comp-vln}, VLMs struggle with spatial reasoning in 4D cities, as indicated by low SR and SPL scores. 
GPT-4o performs best, followed by InternVL3. 
\rev{For comparison, the human reference achieves an SR of 100.0\% and an SPL of 85.87\%, indicating a substantial gap between human and model performance. The average human path length is slightly higher than that of VLMs, mainly because the latter often stop before actually reaching the destination.}
These results, along with recent findings~\cite{DBLP:preprint/arxiv/2407-10943,DBLP:preprint/arxiv/2506-15677}, underscore the difficulty of grounding spatial instructions in complex urban environments and suggest that generated 4D cities can serve as valuable benchmarks for evaluating the spatial reasoning and navigation capabilities of VLMs.

\begin{figure}[!t]
    \centering
    \includegraphics[width=\linewidth]{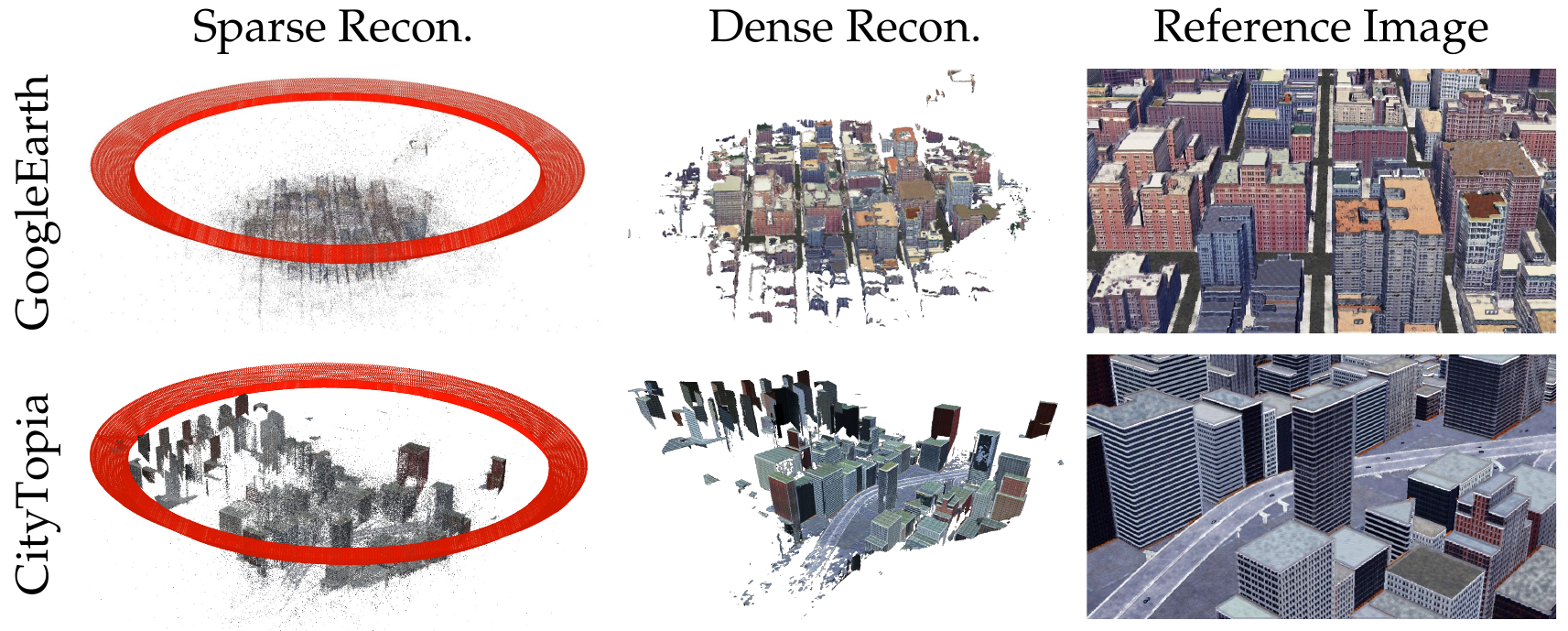}
    \caption{\textbf{COLMAP Reconstruction of 600-frame Orbital Videos.} The red ring shows the camera positions, and the clear point clouds demonstrate \modelname’s consistent rendering. Note that "Recon." stands for "Reconstruction."}
    \label{fig:colmap}
\end{figure}

\begin{figure}[!t]
    \centering
    \includegraphics[width=\linewidth]{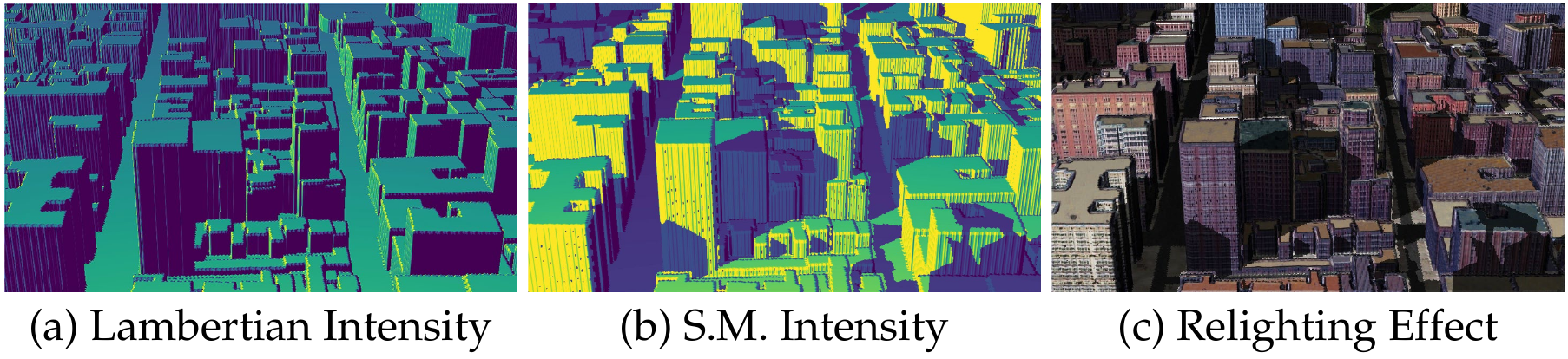}
    \caption{\textbf{Directional Light Relighting Effect.} (a) and (b) show the lighting intensity. (c) illustrates the relighting effect. Note that ``S.M.'' denotes ``Shadow Mapping''.}
    \label{fig:relighting}
\end{figure}

\begin{figure}[!t]
    \centering
    \includegraphics[width=\linewidth]{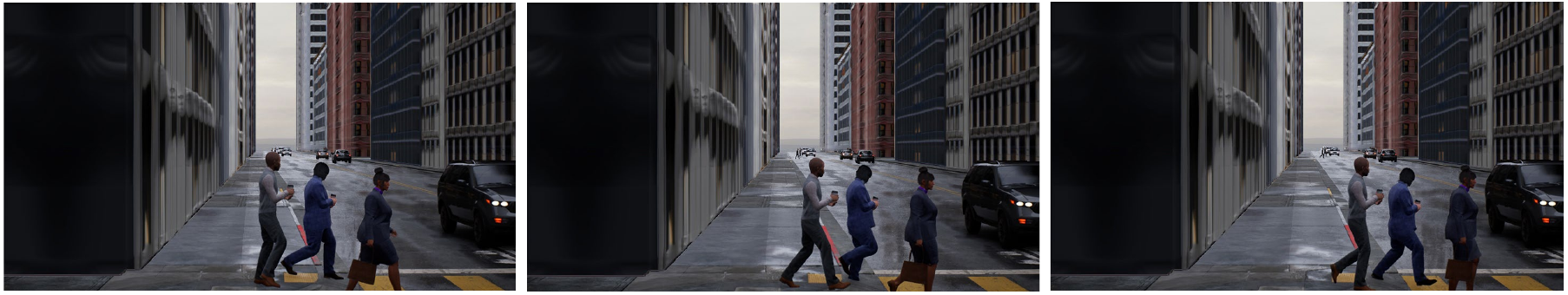}
    \caption{\textbf{Pedestrians in the generated 4D cities.} The three consecutive frames illustrate a group of pedestrians crossing the street.}
    \label{fig:pedestrians}
\end{figure}

\begin{figure}[!t]
    \centering
    \includegraphics[width=\linewidth]{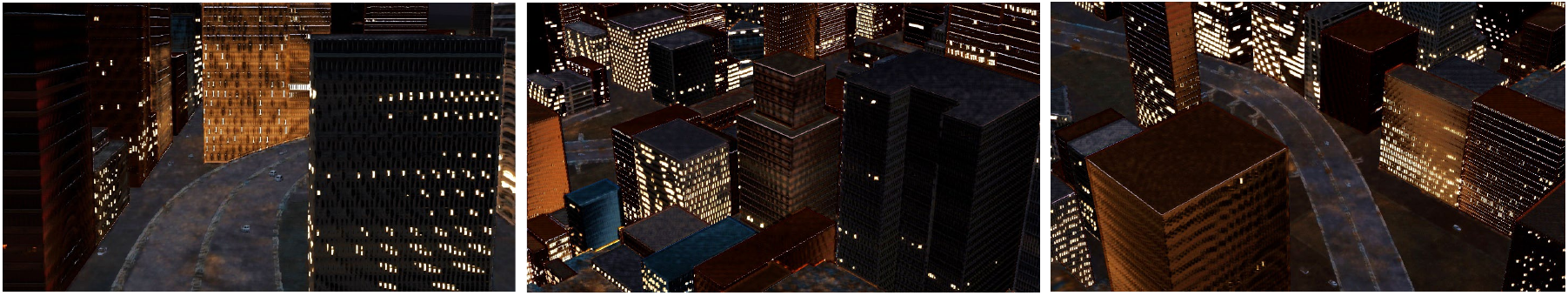}
    \caption{\textbf{Night-view Generation Results.} Despite achieving realistic effects, managing global illumination in the generated scenes remains a challenge.}
    \label{fig:night-views}
\end{figure}

\subsection{Discussions}

\noindent \textbf{View Consistency.}
To demonstrate \modelname's multi-view consistent renderings, we use COLMAP~\cite{DBLP:conf/cvpr/SchonbergerF16} for structure-from-motion and dense reconstruction on orbital videos generated using models trained on the GoogleEarth and CityTopia datasets.
The video sequence comprises 600 frames at a resolution of 960 $\times$ 540, captured from a circular camera trajectory orbiting the scene at a fixed height, with the camera focused on the center. 
Reconstruction is performed solely from the images, without specifying camera parameters. 
As illustrated in Fig.~\ref{fig:colmap}, the estimated camera poses closely align with the sampled trajectory, and the resulting point cloud is both dense and well-defined.

\noindent \textbf{Relighting.}
In \modelname, the generation of background stuff and instances is deliberately decoupled, offering two key benefits: 
\textbf{(1)} Simplified learning for building instances, vehicle instances, and background stuff, and 
\textbf{(2)} Enabling localized editing of building and vehicle instances. 
This approach can be viewed as an inverse rendering process, where \modelname ~generates the albedo, normals, and depth of urban scenes. 
Lighting and shading effects are then computed based on the given lighting conditions.
As shown in Fig.~\ref{fig:relighting}, the shading effects are divided into two components: Lambertian shading and shadow mapping. 
Lambertian shading accounts for the light direction and surface normal, resulting in uniform lighting across all directions, as shown in Fig.~\ref{fig:relighting}\hyperref[fig:relighting]{a}. 
Shadow mapping considers light visibility, enabling the simulation of shadows and occlusion from other objects in the scene, as shown in Fig.~\ref{fig:relighting}\hyperref[fig:relighting]{b}. 
The final relighting effects, with the camera positioned on the left side of the scene, are presented in Fig.~\ref{fig:relighting}\hyperref[fig:relighting]{c}.

\noindent 
\textbf{Diverse Agents Support.}
To explore the potential for supporting more diverse agents in \modelname, we conduct a preliminary experiment integrating pedestrians into the generated scenes. 
We use MoMask~\cite{DBLP:conf/cvpr/GuoMJW024} to synthesize motion, retarget it to 3D human avatars, and render the animated pedestrians into our generated 4D environments.
As shown in Figure~\ref{fig:pedestrians}, the resulting animation demonstrates coherent pedestrian behavior such as street crossing. 
This highlights the feasibility of extending our framework to support richer multi-agent simulations beyond vehicles.

\noindent \textbf{Limitations.}
Despite the realistic generation results, \modelname ~has some limitations.
\textbf{1)} During the inference process, buildings and vehicles are generated individually, leading to a slightly higher computational cost. 
\textbf{2)} The current implementation does not account for global illumination and reflections, which are essential for realistic night scenes.
As illustrated in Fig.~\ref{fig:night-views}, the emitted light from buildings and vehicles does not illuminate the surrounding environment, limiting the realism of the generated cities under such conditions.

\section{Conclusion}

In this paper, we introduce \modelname, a generative model tailored for unbounded 4D city generation. 
Our method simplifies the process by decoupling dynamic objects from static scenes, enabling greater flexibility and realism driven by dynamic traffic scenarios and static city layouts.
Objects in the 4D cities are generated using a composition of stuff-oriented and instance-oriented neural fields for background stuff, buildings, and vehicles.
Additionally, we construct a comprehensive suite of datasets, including OSM, GoogleEarth, and CityTopia, which provide real-world city layouts and cityscapes with high-quality 3D annotations. 
\modelname ~achieves state-of-the-art performance in generating large-scale, realistic 4D cities with instance-level editing, leveraging its compositional design to capture urban diversity and unlock new opportunities for research and practical applications in urban simulation.

\bibliographystyle{IEEEtran}
\bibliography{references}

\begin{thebibliography}{100}
\providecommand{\url}[1]{#1}
\csname url@samestyle\endcsname
\providecommand{\newblock}{\relax}
\providecommand{\bibinfo}[2]{#2}
\providecommand{\BIBentrySTDinterwordspacing}{\spaceskip=0pt\relax}
\providecommand{\BIBentryALTinterwordstretchfactor}{4}
\providecommand{\BIBentryALTinterwordspacing}{\spaceskip=\fontdimen2\font plus
\BIBentryALTinterwordstretchfactor\fontdimen3\font minus \fontdimen4\font\relax}
\providecommand{\BIBforeignlanguage}[2]{{%
\expandafter\ifx\csname l@#1\endcsname\relax
\typeout{** WARNING: IEEEtran.bst: No hyphenation pattern has been}%
\typeout{** loaded for the language `#1'. Using the pattern for}%
\typeout{** the default language instead.}%
\else
\language=\csname l@#1\endcsname
\fi
#2}}
\providecommand{\BIBdecl}{\relax}
\BIBdecl

\bibitem{DBLP:journals/ijcv/XieYZZS20}
H.~Xie, H.~Yao, S.~Zhang, S.~Zhou, and W.~Sun, ``{Pix2Vox++}: Multi-scale context-aware 3{D} object reconstruction from single and multiple images,'' \emph{IJCV}, vol. 128, no.~12, pp. 2919--2935, 2020.

\bibitem{DBLP:conf/iclr/TangRZ0Z24}
J.~Tang, J.~Ren, H.~Zhou, Z.~Liu, and G.~Zeng, ``Dream{G}aussian: Generative {G}aussian splatting for efficient 3{D} content creation,'' in \emph{ICLR}, 2024.

\bibitem{DBLP:preprint/arxiv/2409-12957}
Z.~Chen, J.~Tang, Y.~Dong, Z.~Cao, F.~Hong, Y.~Lan, T.~Wang, H.~Xie, T.~Wu, S.~Saito, L.~Pan, D.~Lin, and Z.~Liu, ``{3DTopia-XL}: High-quality 3{D} {PBR} asset generation via primitive diffusion,'' in \emph{CVPR}, 2025.

\bibitem{DBLP:conf/iclr/Hong0LP023}
F.~Hong, Z.~Chen, Y.~Lan, L.~Pan, and Z.~Liu, ``{EVA3D:} compositional 3{D} human generation from 2{D} image collections,'' in \emph{ICLR}, 2023.

\bibitem{DBLP:conf/nips/0009HMWYL23}
Z.~Chen, F.~Hong, H.~Mei, G.~Wang, L.~Yang, and Z.~Liu, ``Prim{D}iffusion: Volumetric primitives diffusion for 3{D} human generation,'' in \emph{NeurIPS}, 2023.

\bibitem{DBLP:conf/cvpr/LiuZTSZLLL24}
X.~Liu, X.~Zhan, J.~Tang, Y.~Shan, G.~Zeng, D.~Lin, X.~Liu, and Z.~Liu, ``Human{G}aussian: Text-driven 3{D} human generation with {G}aussian splatting,'' in \emph{CVPR}, 2024.

\bibitem{DBLP:journals/pami/ChenWL23}
Z.~Chen, G.~Wang, and Z.~Liu, ``Scene{D}reamer: Unbounded 3{D} scene generation from 2{D} image collections,'' \emph{IEEE TPAMI}, vol.~45, no.~12, pp. 15\,562--15\,576, 2023.

\bibitem{DBLP:journals/tog/WuLYSSWCLSLJ24}
Z.~Wu, Y.~Li, H.~Yan, T.~Shang, W.~Sun, S.~Wang, R.~Cui, W.~Liu, H.~Sato, H.~Li, and P.~Ji, ``Block{F}usion: Expandable 3{D} scene generation using latent tri-plane extrapolation,'' \emph{ACM TOG}, vol.~43, no.~4, pp. 43:1--43:17, 2024.

\bibitem{DBLP:preprint/arxiv/2406-06526}
H.~Xie, Z.~Chen, F.~Hong, and Z.~Liu, ``Generative {G}aussian splatting for unbounded 3{D} city generation,'' in \emph{CVPR}, 2025.

\bibitem{DBLP:conf/iclr/Jiang0GHY24}
Y.~Jiang, L.~Zhang, J.~Gao, W.~Hu, and Y.~Yao, ``Consistent4{D}: Consistent 360{\textdegree} dynamic object generation from monocular video,'' in \emph{ICLR}, 2024.

\bibitem{DBLP:preprint/arxiv/2406-10324}
J.~Ren, K.~Xie, A.~Mirzaei, H.~Liang, X.~Zeng, K.~Kreis, Z.~Liu, A.~Torralba, S.~Fidler, S.~W. Kim, and H.~Ling, ``{L4GM:} large 4{D} {G}aussian reconstruction model,'' in \emph{NeurIPS}, 2024.

\bibitem{DBLP:conf/icml/MaLJFSJ24}
Y.~Ma, Z.~Lin, J.~Ji, Y.~Fan, X.~Sun, and R.~Ji, ``X-{O}scar: {A} progressive framework for high-quality text-guided 3{D} animatable avatar generation,'' in \emph{ICML}, 2024.

\bibitem{DBLP:preprint/arxiv/2406-04629}
Z.~Chai, C.~Tang, Y.~Wong, and M.~S. Kankanhalli, ``{STAR:} skeleton-aware text-based 4{D} avatar generation with in-network motion retargeting,'' \emph{arXiv 2406.04629}, 2024.

\bibitem{DBLP:preprint/arxiv/2407-06188}
X.~Guo, M.~Zhang, H.~Xie, C.~Gu, and Z.~Liu, ``{CrowdMoGen}: Zero-shot text-driven collective motion generation,'' \emph{arXiv 2407.06188}, 2024.

\bibitem{DBLP:conf/iccv/LiuM0SJK21}
A.~Liu, A.~Makadia, R.~Tucker, N.~Snavely, V.~Jampani, and A.~Kanazawa, ``Infinite {N}ature: Perpetual view generation of natural scenes from a single image,'' in \emph{ICCV}, 2021.

\bibitem{DBLP:conf/eccv/LiWSK22}
Z.~Li, Q.~Wang, N.~Snavely, and A.~Kanazawa, ``Infinite{N}ature-{Z}ero: Learning perpetual view generation of natural scenes from single images,'' in \emph{ECCV}, 2022.

\bibitem{DBLP:conf/siggraph/Deng0LGSW24}
B.~Deng, R.~Tucker, Z.~Li, L.~J. Guibas, N.~Snavely, and G.~Wetzstein, ``Streetscapes: Large-scale consistent street view generation using autoregressive video diffusion,'' in \emph{SIGGRAPH}, 2024.

\bibitem{DBLP:conf/cvpr/LiangYLLXC24}
Y.~Liang, X.~Yang, J.~Lin, H.~Li, X.~Xu, and Y.~Chen, ``Lucid{D}reamer: Towards high-fidelity text-to-3{D} generation via interval score matching,'' in \emph{CVPR}, 2024.

\bibitem{DBLP:preprint/arxiv/2404-07199}
J.~Shriram, A.~Trevithick, L.~Liu, and R.~Ramamoorthi, ``Realm{D}reamer: Text-driven 3{D} scene generation with inpainting and depth diffusion,'' in \emph{3DV}, 2025.

\bibitem{DBLP:preprint/arxiv/2406-09394}
H.~Yu, H.~Duan, C.~Herrmann, W.~T. Freeman, and J.~Wu, ``Wonder{W}orld: Interactive 3{D} scene generation from a single image,'' in \emph{CVPR}, 2025.

\bibitem{DBLP:preprint/arxiv/2403-15698}
M.~Zhou, J.~Hou, C.~Luo, Y.~Wang, Z.~Zhang, and J.~Peng, ``Scene{X}: Procedural controllable large-scale scene generation via large-language models,'' in \emph{AAAI}, 2025.

\bibitem{DBLP:preprint/arxiv/abs-2406-04983}
J.~Deng, W.~Chai, J.~Huang, Z.~Zhao, Q.~Huang, M.~Gao, J.~Guo, S.~Hao, W.~Hu, J.~Hwang, X.~Li, and G.~Wang, ``City{C}raft: {A} real crafter for 3{D} city generation,'' \emph{arXiv 2406.04983}, 2024.

\bibitem{DBLP:preprint/arxiv/2407-17572}
S.~Zhang, M.~Zhou, Y.~Wang, C.~Luo, R.~Wang, Y.~Li, X.~Yin, Z.~Zhang, and J.~Peng, ``City{X}: Controllable procedural content generation for unbounded 3{D} cities,'' \emph{arXiv 2407.17572}, 2024.

\bibitem{DBLP:conf/iccv/HaoMB021}
Z.~Hao, A.~Mallya, S.~J. Belongie, and M.~Liu, ``{GANcraft}: Unsupervised 3{D} neural rendering of minecraft worlds,'' in \emph{ICCV}, 2021.

\bibitem{DBLP:conf/cvpr/Park0WZ19}
T.~Park, M.~Liu, T.~Wang, and J.~Zhu, ``Semantic image synthesis with spatially-adaptive normalization,'' in \emph{CVPR}, 2019.

\bibitem{DBLP:conf/iccv/LinLMCS0T23}
C.~H. Lin, H.~Lee, W.~Menapace, M.~Chai, A.~Siarohin, M.~Yang, and S.~Tulyakov, ``Infini{C}ity: Infinite-scale city synthesis,'' in \emph{ICCV}, 2023.

\bibitem{DBLP:conf/cvpr/BahmaniSRWGWTPT24}
S.~Bahmani, I.~Skorokhodov, V.~Rong, G.~Wetzstein, L.~J. Guibas, P.~Wonka, S.~Tulyakov, J.~J. Park, A.~Tagliasacchi, and D.~B. Lindell, ``4{D}-fy: Text-to-4{D} generation using hybrid score distillation sampling,'' in \emph{CVPR}, 2024.

\bibitem{DBLP:conf/cvpr/ZhengLNLHM24}
Y.~Zheng, X.~Li, K.~Nagano, S.~Liu, O.~Hilliges, and S.~D. Mello, ``A unified approach for text-and image-guided 4{D} scene generation,'' in \emph{CVPR}, 2024.

\bibitem{DBLP:preprint/arxiv/2403-16993}
D.~Xu, H.~Liang, N.~P. Bhatt, H.~Hu, H.~Liang, K.~N. Plataniotis, and Z.~Wang, ``Comp4{D}: Llm-guided compositional 4{D} scene generation,'' \emph{arXiv 2403.16993}, 2024.

\bibitem{DBLP:conf/nips/YuWZMSCJTL24}
H.~Yu, C.~Wang, P.~Zhuang, W.~Menapace, A.~Siarohin, J.~Cao, L.~A. Jeni, S.~Tulyakov, and H.~Lee, ``4{R}eal: Towards photorealistic 4{D} scene generation via video diffusion models,'' in \emph{NeurIPS}, 2024.

\bibitem{HAOZHE:link/OpenStreetMap}
\url{https://openstreetmap.org}.

\bibitem{HAOZHE:link/GoogleEarth}
\url{https://earth.google.com/studio}.

\bibitem{HAOZHE:link/CitySample}
\url{https://www.unrealengine.com/marketplace/en-US/product/city-sample}.

\bibitem{DBLP:conf/cvpr/XieCHL24}
H.~Xie, Z.~Chen, F.~Hong, and Z.~Liu, ``City{D}reamer: Compositional generative model of unbounded 3{D} cities,'' in \emph{CVPR}, 2024.

\bibitem{DBLP:journals/pami/KarrasLA21}
T.~Karras, S.~Laine, and T.~Aila, ``A style-based generator architecture for generative adversarial networks,'' \emph{IEEE TPAMI}, vol.~43, no.~12, pp. 4217--4228, 2021.

\bibitem{DBLP:journals/pami/MelnikMMPAHRRR24}
A.~Melnik, M.~Miasayedzenkau, D.~Makarovets, D.~Pirshtuk, E.~Akbulut, D.~Holzmann, T.~Renusch, G.~Reichert, and H.~J. Ritter, ``Face generation and editing with {StyleGAN}: {A} survey,'' \emph{IEEE TPAMI}, vol.~46, no.~5, pp. 3557--3576, 2024.

\bibitem{DBLP:conf/nips/GoodfellowPMXWOCB14}
I.~J. Goodfellow, J.~Pouget{-}Abadie, M.~Mirza, B.~Xu, D.~Warde{-}Farley, S.~Ozair, A.~C. Courville, and Y.~Bengio, ``Generative adversarial nets,'' in \emph{NIPS}, 2014.

\bibitem{DBLP:conf/3dim/GadelhaMW17}
M.~Gadelha, S.~Maji, and R.~Wang, ``3{D} shape induction from 2{D} views of multiple objects,'' in \emph{3DV}, 2017.

\bibitem{DBLP:conf/iccv/HenzlerM019}
P.~Henzler, N.~J. Mitra, and T.~Ritschel, ``Escaping plato's cave: 3{D} shape from adversarial rendering,'' in \emph{ICCV}, 2019.

\bibitem{DBLP:conf/iccv/Nguyen-PhuocLTR19}
T.~Nguyen{-}Phuoc, C.~Li, L.~Theis, C.~Richardt, and Y.~Yang, ``Hologan: Unsupervised learning of 3d representations from natural images,'' in \emph{ICCV}, 2019.

\bibitem{DBLP:preprint/arxiv/1910-00287}
A.~Szab{\'{o}}, G.~Meishvili, and P.~Favaro, ``Unsupervised generative 3{D} shape learning from natural images,'' \emph{arXiv 1910.00287}, 2019.

\bibitem{DBLP:conf/cvpr/LiaoSMG20}
Y.~Liao, K.~Schwarz, L.~M. Mescheder, and A.~Geiger, ``Towards unsupervised learning of generative models for 3{D} controllable image synthesis,'' in \emph{CVPR}, 2020.

\bibitem{DBLP:conf/eccv/MildenhallSTBRN20}
B.~Mildenhall, P.~P. Srinivasan, M.~Tancik, J.~T. Barron, R.~Ramamoorthi, and R.~Ng, ``{NeRF}: Representing scenes as neural radiance fields for view synthesis,'' in \emph{ECCV}, 2020.

\bibitem{DBLP:conf/nips/SchwarzLN020}
K.~Schwarz, Y.~Liao, M.~Niemeyer, and A.~Geiger, ``{GRAF:} generative radiance fields for 3{D}-aware image synthesis,'' in \emph{NeurIPS}, 2020.

\bibitem{DBLP:conf/cvpr/ChanMK0W21}
E.~R. Chan, M.~Monteiro, P.~Kellnhofer, J.~Wu, and G.~Wetzstein, ``Pi-{GAN}: Periodic implicit generative adversarial networks for 3{D}-aware image synthesis,'' in \emph{CVPR}, 2021.

\bibitem{DBLP:conf/iccv/DeVries0STS21}
T.~DeVries, M.~{\'{A}}. Bautista, N.~Srivastava, G.~W. Taylor, and J.~M. Susskind, ``Unconstrained scene generation with locally conditioned radiance fields,'' in \emph{ICCV}, 2021.

\bibitem{DBLP:conf/nips/XuPLD21}
X.~Xu, X.~Pan, D.~Lin, and B.~Dai, ``Generative occupancy fields for 3d surface-aware image synthesis,'' in \emph{NeurIPS}, 2021.

\bibitem{DBLP:conf/cvpr/Niemeyer021}
M.~Niemeyer and A.~Geiger, ``{GIRAFFE:} representing scenes as compositional generative neural feature fields,'' in \emph{CVPR}, 2021.

\bibitem{DBLP:conf/iclr/GuL0T22}
J.~Gu, L.~Liu, P.~Wang, and C.~Theobalt, ``{StyleNeRF}: {A} style-based 3{D} aware generator for high-resolution image synthesis,'' in \emph{ICLR}, 2022.

\bibitem{DBLP:conf/cvpr/XueLSL22}
Y.~Xue, Y.~Li, K.~K. Singh, and Y.~J. Lee, ``{GIRAFFE} {HD:} {A} high-resolution 3{D}-aware generative model,'' in \emph{CVPR}, 2022.

\bibitem{DBLP:conf/cvpr/Or-ElLSSPK22}
R.~Or{-}El, X.~Luo, M.~Shan, E.~Shechtman, J.~J. Park, and I.~Kemelmacher{-}Shlizerman, ``{StyleSDF}: High-resolution 3{D}-consistent image and geometry generation,'' in \emph{CVPR}, 2022.

\bibitem{DBLP:conf/cvpr/ChanLCNPMGGTKKW22}
E.~R. Chan, C.~Z. Lin, M.~A. Chan, K.~Nagano, B.~Pan, S.~D. Mello, O.~Gallo, L.~J. Guibas, J.~Tremblay, S.~Khamis, T.~Karras, and G.~Wetzstein, ``Efficient geometry-aware 3d generative adversarial networks,'' in \emph{CVPR}, 2022.

\bibitem{DBLP:conf/nips/SchwarzSNL022}
K.~Schwarz, A.~Sauer, M.~Niemeyer, Y.~Liao, and A.~Geiger, ``{VoxGRAF}: Fast 3{D}-aware image synthesis with sparse voxel grids,'' in \emph{NeurIPS}, 2022.

\bibitem{DBLP:conf/cvpr/DengYX022}
Y.~Deng, J.~Yang, J.~Xiang, and X.~Tong, ``{GRAM:} generative radiance manifolds for 3{D}-aware image generation,'' in \emph{CVPR}, 2022.

\bibitem{DBLP:conf/eccv/ZhaoMGRSC22}
X.~Zhao, F.~Ma, D.~G{\"{u}}era, Z.~Ren, A.~G. Schwing, and A.~Colburn, ``Generative multiplane images: Making a 2{D} {GAN} 3{D}-aware,'' in \emph{ECCV}, 2022.

\bibitem{DBLP:conf/cvpr/KarrasLA19}
T.~Karras, S.~Laine, and T.~Aila, ``A style-based generator architecture for generative adversarial networks,'' in \emph{CVPR}, 2019.

\bibitem{DBLP:conf/cvpr/Yang0WHSYC20}
H.~Yang, H.~Zhu, Y.~Wang, M.~Huang, Q.~Shen, R.~Yang, and X.~Cao, ``Face{S}cape: {A} large-scale high quality 3{D} face dataset and detailed riggable 3d face prediction,'' in \emph{CVPR}, 2020.

\bibitem{DBLP:journals/pami/IonescuPOS14}
C.~Ionescu, D.~Papava, V.~Olaru, and C.~Sminchisescu, ``Human3.6{M}: Large scale datasets and predictive methods for 3{D} human sensing in natural environments,'' \emph{IEEE TPAMI}, vol.~36, no.~7, pp. 1325--1339, 2014.

\bibitem{DBLP:conf/eccv/CaiRZLYWFGYPHZL22}
Z.~Cai, D.~Ren, A.~Zeng, Z.~Lin, T.~Yu, W.~Wang, X.~Fan, Y.~Gao, Y.~Yu, L.~Pan, F.~Hong, M.~Zhang, C.~C. Loy, L.~Yang, and Z.~Liu, ``{HuMMan}: Multi-modal 4{D} human dataset for versatile sensing and modeling,'' in \emph{ECCV}, 2022.

\bibitem{DBLP:conf/cvpr/WuZFWRPWYWQLL23}
T.~Wu, J.~Zhang, X.~Fu, Y.~Wang, J.~Ren, L.~Pan, W.~Wu, L.~Yang, J.~Wang, C.~Qian, D.~Lin, and Z.~Liu, ``{OmniObject3D}: Large-vocabulary 3{D} object dataset for realistic perception, reconstruction and generation,'' in \emph{CVPR}, 2023.

\bibitem{DBLP:conf/cvpr/DeitkeSSWMVSEKF23}
M.~Deitke, D.~Schwenk, J.~Salvador, L.~Weihs, O.~Michel, E.~VanderBilt, L.~Schmidt, K.~Ehsani, A.~Kembhavi, and A.~Farhadi, ``Objaverse: {A} universe of annotated 3{D} objects,'' in \emph{CVPR}, 2023.

\bibitem{DBLP:preprint/arxiv/2505-05474}
B.~Wen, H.~Xie, Z.~Chen, F.~Hong, and Z.~Liu, ``3d scene generation: A survey,'' \emph{arXiv 2505.05474}, 2025.

\bibitem{DBLP:journals/pami/ZhanYWZLLKTX23}
F.~Zhan, Y.~Yu, R.~Wu, J.~Zhang, S.~Lu, L.~Liu, A.~Kortylewski, C.~Theobalt, and E.~P. Xing, ``Multimodal image synthesis and editing: The generative {AI} era,'' \emph{IEEE TPAMI}, vol.~45, no.~12, pp. 15\,098--15\,119, 2023.

\bibitem{DBLP:conf/cvpr/RaistrickLMMWZK23}
A.~Raistrick, L.~Lipson, Z.~Ma, L.~Mei, M.~Wang, Y.~Zuo, K.~Kayan, H.~Wen, B.~Han, Y.~Wang, A.~Newell, H.~Law, A.~Goyal, K.~Yang, and J.~Deng, ``Infinite photorealistic worlds using procedural generation,'' in \emph{CVPR}, 2023.

\bibitem{DBLP:journals/tog/LiPXCKSTCCZ19}
M.~Li, A.~G. Patil, K.~Xu, S.~Chaudhuri, O.~Khan, A.~Shamir, C.~Tu, B.~Chen, D.~Cohen{-}Or, and H.~R. Zhang, ``{GRAINS:} generative recursive autoencoders for indoor scenes,'' \emph{ACM TOG}, vol.~38, no.~2, pp. 12:1--12:16, 2019.

\bibitem{DBLP:conf/iccv/FuC0ZWLZSJZ021}
H.~Fu, B.~Cai, L.~Gao, L.~Zhang, J.~Wang, C.~Li, Q.~Zeng, C.~Sun, R.~Jia, B.~Zhao, and H.~Zhang, ``{3D-FRONT}: 3{D} furnished rooms with layouts and semantics,'' in \emph{ICCV}, 2021.

\bibitem{DBLP:conf/cvpr/RaistrickMKY0HW24}
A.~Raistrick, L.~Mei, K.~Kayan, D.~Yan, Y.~Zuo, B.~Han, H.~Wen, M.~Parakh, S.~Alexandropoulos, L.~Lipson, Z.~Ma, and J.~Deng, ``Infinigen {I}ndoors: Photorealistic indoor scenes using procedural generation,'' in \emph{CVPR}, 2024.

\bibitem{DBLP:conf/corl/DosovitskiyRCLK17}
A.~Dosovitskiy, G.~Ros, F.~Codevilla, A.~M. L{\'{o}}pez, and V.~Koltun, ``{CARLA:} an open urban driving simulator,'' in \emph{CoRL}, 2017.

\bibitem{DBLP:preprint/arxiv/1909-11512}
S.~I. Nikolenko, ``Synthetic data for deep learning,'' \emph{arXiv 1909.11512}, 2019.

\bibitem{DBLP:preprint/arxiv/2412-07660}
Y.~Li, X.~Ran, L.~Xu, T.~Lu, M.~Yu, Z.~Wang, Y.~Xiangli, D.~Lin, and B.~Dai, ``Proc-{GS}: Procedural building generation for city assembly with 3{D} {G}aussians,'' in \emph{CVPR Workshops}, 2025.

\bibitem{DBLP:journals/pami/JiangKS16}
Y.~Jiang, H.~S. Koppula, and A.~Saxena, ``Modeling 3{D} environments through hidden human context,'' \emph{IEEE TPAMI}, vol.~38, no.~10, pp. 2040--2053, 2016.

\bibitem{DBLP:conf/nips/PaschalidouKSKG21}
D.~Paschalidou, A.~Kar, M.~Shugrina, K.~Kreis, A.~Geiger, and S.~Fidler, ``{ATISS:} autoregressive transformers for indoor scene synthesis,'' in \emph{NeurIPS}, 2021.

\bibitem{DBLP:journals/pami/GaoSMLGY23}
L.~Gao, J.~Sun, K.~Mo, Y.~Lai, L.~J. Guibas, and J.~Yang, ``Scene{HGN}: Hierarchical graph networks for 3{D} indoor scene generation with fine-grained geometry,'' \emph{IEEE TPAMI}, vol.~45, no.~7, pp. 8902--8919, 2023.

\bibitem{DBLP:conf/cvpr/DaiCSHFN17}
A.~Dai, A.~X. Chang, M.~Savva, M.~Halber, T.~A. Funkhouser, and M.~Nie{\ss}ner, ``Scan{N}et: Richly-annotated 3{D} reconstructions of indoor scenes,'' in \emph{CVPR}, 2017.

\bibitem{DBLP:preprint/arxiv/1906-05797}
J.~Straub, T.~Whelan, L.~Ma, Y.~Chen, E.~Wijmans, S.~Green, J.~J. Engel, R.~Mur{-}Artal, C.~Y. Ren, S.~Verma, A.~Clarkson, M.~Yan, B.~Budge, Y.~Yan, X.~Pan, J.~Yon, Y.~Zou, K.~Leon, N.~Carter, J.~Briales, T.~Gillingham, E.~Mueggler, L.~Pesqueira, M.~Savva, D.~Batra, H.~M. Strasdat, R.~{De Nardi}, M.~Goesele, S.~Lovegrove, and R.~A. Newcombe, ``The {R}eplica {D}ataset: {A} digital replica of indoor spaces,'' \emph{arXiv 1906.05797}, 2019.

\bibitem{DBLP:conf/cvpr/SongYZCSF17}
S.~Song, F.~Yu, A.~Zeng, A.~X. Chang, M.~Savva, and T.~A. Funkhouser, ``Semantic scene completion from a single depth image,'' in \emph{CVPR}, 2017.

\bibitem{DBLP:journals/ijcv/FuJGGZMT21}
H.~Fu, R.~Jia, L.~Gao, M.~Gong, B.~Zhao, S.~J. Maybank, and D.~Tao, ``{3D-FUTURE}: 3{D} furniture shape with texture,'' \emph{IJCV}, vol. 129, no.~12, pp. 3313--3337, 2021.

\bibitem{DBLP:conf/corl/DaiWJWGZWF24}
T.~Dai, J.~Wong, Y.~Jiang, C.~Wang, C.~Gokmen, R.~Zhang, J.~Wu, and L.~Fei-Fei, ``{ACDC}: Automated creation of digital cousins for robust policy learning,'' in \emph{CoRL}, 2024.

\bibitem{DBLP:conf/cvpr/PumarolaCPM21}
A.~Pumarola, E.~Corona, G.~Pons{-}Moll, and F.~Moreno{-}Noguer, ``{D-NeRF}: Neural radiance fields for dynamic scenes,'' in \emph{CVPR}, 2021.

\bibitem{DBLP:conf/cvpr/YangGZJ0024}
Z.~Yang, X.~Gao, W.~Zhou, S.~Jiao, Y.~Zhang, and X.~Jin, ``Deformable 3{D} {G}aussians for high-fidelity monocular dynamic scene reconstruction,'' in \emph{CVPR}, 2024.

\bibitem{DBLP:preprint/arxiv/2410-10429}
S.~Gu, W.~Yin, B.~Jin, X.~Guo, J.~Wang, H.~Li, Q.~Zhang, and X.~Long, ``{DOME}: Taming diffusion model into high-fidelity controllable occupancy world model,'' \emph{arXiv 2410.10429}, 2024.

\bibitem{DBLP:preprint/arxiv/2410-18084}
H.~Bian, L.~Kong, H.~Xie, L.~Pan, Y.~Qiao, and Z.~Liu, ``Dynamic{C}ity: Large-scale {LiDAR} generation from dynamic scenes,'' in \emph{ICLR}, 2025.

\bibitem{DBLP:conf/cvpr/ChangZJLF22}
H.~Chang, H.~Zhang, L.~Jiang, C.~Liu, and W.~T. Freeman, ``Mask{GIT}: Masked generative image transformer,'' in \emph{CVPR}, 2022.

\bibitem{DBLP:conf/nips/OordVK17}
A.~van~den Oord, O.~Vinyals, and K.~Kavukcuoglu, ``Neural discrete representation learning,'' in \emph{NIPS}, 2017.

\bibitem{DBLP:conf/aaai/MeisterH018}
S.~Meister, J.~Hur, and S.~Roth, ``Un{F}low: Unsupervised learning of optical flow with a bidirectional census loss,'' in \emph{AAAI}, 2018.

\bibitem{DBLP:conf/eccv/Bond-TaylorH0BW22}
S.~Bond{-}Taylor, P.~Hessey, H.~Sasaki, T.~P. Breckon, and C.~G. Willcocks, ``Unleashing transformers: Parallel token prediction with discrete absorbing diffusion for fast high-resolution image generation from vector-quantized codes,'' in \emph{ECCV}, 2022.

\bibitem{DBLP:conf/icra/FengLPTZ23}
L.~Feng, Q.~Li, Z.~Peng, S.~Tan, and B.~Zhou, ``Traffic{G}en: Learning to generate diverse and realistic traffic scenarios,'' in \emph{ICRA}, 2023.

\bibitem{DBLP:conf/iccv/EttingerCCLZPCS21}
S.~Ettinger, S.~Cheng, B.~Caine, C.~Liu, H.~Zhao, S.~Pradhan, Y.~Chai, B.~Sapp, C.~R. Qi, Y.~Zhou, Z.~Yang, A.~Chouard, P.~Sun, J.~Ngiam, V.~Vasudevan, A.~McCauley, J.~Shlens, and D.~Anguelov, ``Large scale interactive motion forecasting for autonomous driving : The {W}aymo open motion dataset,'' in \emph{ICCV}, 2021.

\bibitem{DBLP:journals/pami/Canny86a}
J.~F. Canny, ``A computational approach to edge detection,'' \emph{IEEE TPAMI}, vol.~8, no.~6, pp. 679--698, 1986.

\bibitem{DBLP:journals/cacm/ZhangS84}
T.~Y. Zhang and C.~Y. Suen, ``A fast parallel algorithm for thinning digital patterns,'' \emph{Communications of the {ACM}}, vol.~27, no.~3, pp. 236--239, 1984.

\bibitem{DBLP:conf/eccv/JohnsonAF16}
J.~Johnson, A.~Alahi, and L.~Fei{-}Fei, ``Perceptual losses for real-time style transfer and super-resolution,'' in \emph{ECCV}, 2016.

\bibitem{DBLP:arxiv/LimY17}
J.~H. Lim and J.~C. Ye, ``Geometric {GAN},'' \emph{arXiv 1705.02894}, 2017.

\bibitem{DBLP:conf/cvpr/GeigerLU12}
A.~Geiger, P.~Lenz, and R.~Urtasun, ``Are we ready for autonomous driving? the {KITTI} vision benchmark suite,'' in \emph{CVPR}, 2012.

\bibitem{DBLP:conf/cvpr/CordtsORREBFRS16}
M.~Cordts, M.~Omran, S.~Ramos, T.~Rehfeld, M.~Enzweiler, R.~Benenson, U.~Franke, S.~Roth, and B.~Schiele, ``The {C}ityscapes dataset for semantic urban scene understanding,'' in \emph{CVPR}, 2016.

\bibitem{DBLP:conf/wacv/NigamHR18}
I.~Nigam, C.~Huang, and D.~Ramanan, ``Ensemble knowledge transfer for semantic segmentation,'' in \emph{WACV}, 2018.

\bibitem{DBLP:conf/cvpr/CaesarBLVLXKPBB20}
H.~Caesar, V.~Bankiti, A.~H. Lang, S.~Vora, V.~E. Liong, Q.~Xu, A.~Krishnan, Y.~Pan, G.~Baldan, and O.~Beijbom, ``nu{S}cenes: {A} multimodal dataset for autonomous driving,'' in \emph{CVPR}, 2020.

\bibitem{DBLP:conf/eccv/RichterVRK16}
S.~R. Richter, V.~Vineet, S.~Roth, and V.~Koltun, ``Playing for data: Ground truth from computer games,'' in \emph{ECCV}, 2016.

\bibitem{DBLP:conf/cvpr/RosSMVL16}
G.~Ros, L.~Sellart, J.~Materzynska, D.~V{\'{a}}zquez, and A.~M. L{\'{o}}pez, ``The {SYNTHIA} dataset: {A} large collection of synthetic images for semantic segmentation of urban scenes,'' in \emph{CVPR}, 2016.

\bibitem{DBLP:conf/eccv/SalehASPA18}
F.~S. Saleh, M.~S. Aliakbarian, M.~Salzmann, L.~Petersson, and J.~M. {\'{A}}lvarez, ``Effective use of synthetic data for urban scene semantic segmentation,'' in \emph{ECCV}, 2018.

\bibitem{DBLP:conf/iccv/0002JXX0L023}
Y.~Li, L.~Jiang, L.~Xu, Y.~Xiangli, Z.~Wang, D.~Lin, and B.~Dai, ``Matrix{C}ity: {A} large-scale city dataset for city-scale neural rendering and beyond,'' in \emph{ICCV}, 2023.

\bibitem{DBLP:preprint/arxiv/2008-03286}
Y.~Zhou, J.~Huang, X.~Dai, L.~Luo, Z.~Chen, and Y.~Ma, ``Holi{C}ity: {A} city-scale data platform for learning holistic 3{D} structures,'' \emph{arXiv 2008.03286}, 2020.

\bibitem{DBLP:journals/pami/LiaoXG23}
Y.~Liao, J.~Xie, and A.~Geiger, ``{KITTI-360:} {A} novel dataset and benchmarks for urban scene understanding in 2{D} and 3{D},'' \emph{IEEE TPAMI}, vol.~45, no.~3, pp. 3292--3310, 2023.

\bibitem{DBLP:conf/eccv/LinLHYXH22}
L.~Lin, Y.~Liu, Y.~Hu, X.~Yan, K.~Xie, and H.~Huang, ``Capturing, reconstructing, and simulating: The {UrbanScene3D} dataset,'' in \emph{ECCV}, 2022.

\bibitem{DBLP:conf/siggraph/Perlin85}
K.~Perlin, ``An image synthesizer,'' in \emph{SIGGRAPH}, 1985.

\bibitem{DBLP:conf/nips/ShenMW22}
Y.~Shen, W.~Ma, and S.~Wang, ``{SGAM:} building a virtual 3{D} world through simultaneous generation and mapping,'' in \emph{NeurIPS}, 2022.

\bibitem{DBLP:conf/cvpr/Chai0LIS23}
L.~Chai, R.~Tucker, Z.~Li, P.~Isola, and N.~Snavely, ``Persistent {N}ature: {A} generative model of unbounded 3{D} worlds,'' in \emph{CVPR}, 2023.

\bibitem{DBLP:conf/nips/ChuKF24}
W.~Chu, L.~Ke, and K.~Fragkiadaki, ``Dreamscene4{D}: Dynamic multi-object scene generation from monocular videos,'' in \emph{NeurIPS}, 2024.

\bibitem{DBLP:preprint/arxiv/2411-04928}
W.~Sun, S.~Chen, F.~Liu, Z.~Chen, Y.~Duan, J.~Zhang, and Y.~Wang, ``Dimension{X}: Create any 3{D} and 4{D} scenes from a single image with controllable video diffusion,'' in \emph{ICCV}, 2025.

\bibitem{DBLP:conf/nips/HeuselRUNH17}
M.~Heusel, H.~Ramsauer, T.~Unterthiner, B.~Nessler, and S.~Hochreiter, ``{GANs} trained by a two time-scale update rule converge to a local nash equilibrium,'' in \emph{NIPS}, 2017.

\bibitem{DBLP:conf/iclr/BinkowskiSAG18}
M.~Binkowski, D.~J. Sutherland, M.~Arbel, and A.~Gretton, ``Demystifying {MMD} {GAN}s,'' in \emph{ICLR}, 2018.

\bibitem{DBLP:conf/cvpr/HuangHYZS0Z0JCW24}
Z.~Huang, Y.~He, J.~Yu, F.~Zhang, C.~Si, Y.~Jiang, Y.~Zhang, T.~Wu, Q.~Jin, N.~Chanpaisit, Y.~Wang, X.~Chen, L.~Wang, D.~Lin, Y.~Qiao, and Z.~Liu, ``V{B}ench: Comprehensive benchmark suite for video generative models,'' in \emph{CVPR}, 2024.

\bibitem{DBLP:journals/pami/RanftlLHSK22}
R.~Ranftl, K.~Lasinger, D.~Hafner, K.~Schindler, and V.~Koltun, ``Towards robust monocular depth estimation: Mixing datasets for zero-shot cross-dataset transfer,'' \emph{IEEE TPAMI}, vol.~44, no.~3, pp. 1623--1637, 2022.

\bibitem{DBLP:conf/cvpr/SchonbergerF16}
J.~L. Sch{\"{o}}nberger and J.~Frahm, ``Structure-from-motion revisited,'' in \emph{CVPR}, 2016.

\bibitem{DBLP:journals/tog/ChenEWMZ08}
G.~Chen, G.~Esch, P.~Wonka, P.~M{\"{u}}ller, and E.~Zhang, ``Interactive procedural street modeling,'' \emph{ACM TOG}, vol.~27, no.~3, p. 103, 2008.

\bibitem{DBLP:conf/iclr/LinLCT022}
C.~H. Lin, H.~Lee, Y.~Cheng, S.~Tulyakov, and M.~Yang, ``Infinity{G}an: Towards infinite-pixel image synthesis,'' in \emph{ICLR}, 2022.

\bibitem{DBLP:conf/iccv/ZhangRA23}
L.~Zhang, A.~Rao, and M.~Agrawala, ``Adding conditional control to text-to-image diffusion models,'' in \emph{ICCV}, 2023.

\bibitem{DBLP:preprint/arxiv/2312-11805}
R.~Anil, S.~Borgeaud, Y.~Wu, J.~Alayrac, J.~Yu, R.~Soricut, J.~Schalkwyk, A.~M. Dai, A.~Hauth, K.~Millican, D.~Silver, S.~Petrov, M.~Johnson, I.~Antonoglou, J.~Schrittwieser, A.~Glaese, J.~Chen, E.~Pitler, T.~P. Lillicrap, A.~Lazaridou, and et~al., ``Gemini: {A} family of highly capable multimodal models,'' \emph{arXiv 2312.11805}, 2023.

\bibitem{DBLP:preprint/arxiv/2410-21276}
OpenAI, ``{GPT}-4o system card,'' \emph{arXiv 2410.21276}, 2024.

\bibitem{DBLP:conf/acl/DongKYLFR25}
H.~Dong, Z.~Kang, W.~Yin, X.~Liang, C.~Feng, and J.~Ran, ``Scalable vision language model training via high quality data curation,'' in \emph{ACL}, 2025.

\bibitem{DBLP:preprint/arxiv/2405-20797}
S.~Lu, Y.~Li, Q.~Chen, Z.~Xu, W.~Luo, K.~Zhang, and H.~Ye, ``Ovis: Structural embedding alignment for multimodal large language model,'' \emph{arXiv 2405.20797}, 2024.

\bibitem{DBLP:preprint/arxiv/2502-13923}
S.~Bai, K.~Chen, X.~Liu, J.~Wang, W.~Ge, S.~Song, K.~Dang, P.~Wang, S.~Wang, J.~Tang, H.~Zhong, Y.~Zhu, M.~Yang, Z.~Li, J.~Wan, P.~Wang, W.~Ding, Z.~Fu, Y.~Xu, J.~Ye, X.~Zhang, T.~Xie, Z.~Cheng, H.~Zhang, Z.~Yang, H.~Xu, and J.~Lin, ``{Qwen2.5-VL} technical report,'' \emph{arXiv 2502.13923}, 2025.

\bibitem{DBLP:preprint/arxiv/2502-04328}
Z.~Liu, Y.~Dong, J.~Wang, Z.~Liu, W.~Hu, J.~Lu, and Y.~Rao, ``Ola: Pushing the frontiers of omni-modal language model with progressive modality alignment,'' \emph{arXiv 2502.04328}, 2025.

\bibitem{DBLP:preprint/arxiv/2504-10479}
J.~Zhu, W.~Wang, Z.~Chen, Z.~Liu, S.~Ye, L.~Gu, H.~Tian, Y.~Duan, W.~Su, J.~Shao, Z.~Gao, E.~Cui, X.~Wang, Y.~Cao, Y.~Liu, X.~Wei, H.~Zhang, H.~Wang, W.~Xu, H.~Li, and et~al., ``{InternVL3:} exploring advanced training and test-time recipes for open-source multimodal models,'' \emph{arXiv 2504.10479}, 2025.

\bibitem{DBLP:preprint/arxiv/2407-10943}
H.~Wang, J.~Chen, W.~Huang, Q.~Ben, T.~Wang, B.~Mi, T.~Huang, S.~Zhao, Y.~Chen, S.~Yang, P.~Cao, W.~Yu, Z.~Ye, J.~Li, J.~Long, Z.~Wang, H.~Wang, Y.~Zhao, Z.~Tu, Y.~Qiao, D.~Lin, and J.~Pang, ``{GRUtopia:} dream general robots in a city at scale,'' \emph{arXiv 2407.10943}, 2024.

\bibitem{DBLP:preprint/arxiv/2506-15677}
Y.~Hong, R.~Sun, B.~Li, X.~Yao, M.~Wu, A.~Chien, D.~Yin, Y.~N. Wu, Z.~J. Wang, and K.-W. Chang, ``Embodied {W}eb {A}gents: Bridging physical-digital realms for integrated agent intelligence,'' \emph{arXiv 2506.15677}, 2025.

\bibitem{DBLP:conf/cvpr/GuoMJW024}
C.~Guo, Y.~Mu, M.~G. Javed, S.~Wang, and L.~Cheng, ``{MoMask}: Generative masked modeling of 3{D} human motions,'' in \emph{CVPR}, 2024.

\end{thebibliography}


\end{document}